\pdfoutput=1
\documentclass[11pt]{article}

\usepackage[final]{acl}

\usepackage{times}
\usepackage{latexsym}

\usepackage[T1]{fontenc}
\usepackage[utf8]{inputenc}

\usepackage{microtype}

\usepackage{inconsolata}

\usepackage{graphicx}
\usepackage{longtable}

\usepackage{threeparttable}
\usepackage{booktabs}
\usepackage{multirow}
\usepackage{makecell}
\title{CliniChat: A Multi-Source Knowledge-Driven Framework for Clinical Interview Dialogue Reconstruction and Evaluation}

\author{
  Jing Chen$^{*1}$, Zhihua Wei$^{*1}$, Wei Zhang$^2$, Yingying Hu$^2$, Qiong Zhang$^3$\\
  $^1$College of Computer Science and Technology, Tongji University, Shanghai, China\\
  $^2$Zhongnan Hospital of Wuhan University, Wuhan, China\\
  $^3$Renmin Hospital of Wuhan University, Wuhan, China\\
  \{chenjing\_miss,zhihua\_wei\}@tongji.edu.cn
}

\begin{document}
\maketitle
\begin{abstract}
Large language models (LLMs) hold great promise for assisting clinical interviews due to their fluent interactive capabilities and extensive medical knowledge. However, the lack of high-quality interview dialogue data and widely accepted evaluation methods has significantly impeded this process. So we propose CliniChat, a framework that integrates multi-source knowledge to enable LLMs to simulate real-world clinical interviews. It consists of two modules: Clini-Recon and Clini-Eval, each responsible for reconstructing and evaluating interview dialogues, respectively. By incorporating three sources of knowledge, Clini-Recon transforms clinical notes into systematic, professional, and empathetic interview dialogues. Clini-Eval combines a comprehensive evaluation metric system with a two-phase automatic evaluation approach, enabling LLMs to assess interview performance like experts. We contribute MedQA-Dialog, a high-quality synthetic interview dialogue dataset, and CliniChatGLM, a model specialized for clinical interviews. Experimental results demonstrate that CliniChatGLM\textquotesingle s interview capabilities undergo a comprehensive upgrade, particularly in history-taking, achieving state-of-the-art performance.
\end{abstract}

\section{Introduction}
The clinical interview is the most fundamental task performed by physicians, spanning from history taking and physical examination to preliminary diagnosis. It involves intensive physician-patient interaction, especially during history taking, when physicians must inquire in detail with patients or their families to fully grasp the patient's medical history~\citep{butler2023history}. Research has shown that physicians can reach a final diagnosis for 76\% of cases based solely on good history taking~\citep{keifenheim2015teaching}. For physicians, the clinical interview is a time-consuming and knowledge-intensive medical practice. A satisfactory interview often requires up to 40 rounds of physician-patient interaction~\citep{CHENZi-xuan2023MedicineandPhilosophy}, during which the physician must not only follow a structured interview process but also skillfully apply expertise, interview techniques, and diagnostic reasoning.

\begin{figure}[t]
  \includegraphics[width=\columnwidth]{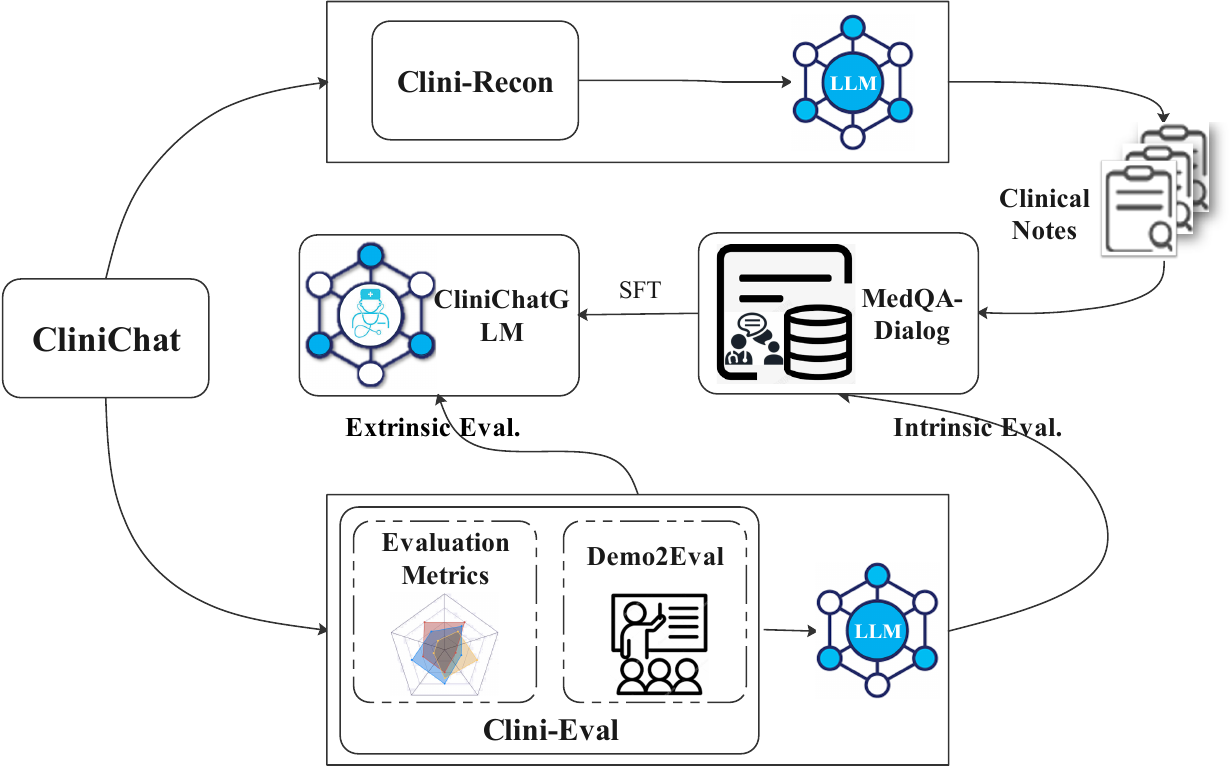}
  \caption{An overview of the CliniChat framework.}
  \label{fig1}
\end{figure}

For a long time, the NLP community has been committed to developing tools to assist physicians in clinical interviews~\citep{nash2010isabel,chung2019chatbot,9289621}. However, due to limited knowledge coverage and interaction capabilities, these tools have failed to gain widespread adoption. With the advent of large language models (LLMs), the field of assisted clinical interviews has been revitalized. Specifically, LLMs offer two key advantages: they support smooth human-machine interaction, and some LLMs~\citep{achiam2023gpt, erniebot2024, singhal2023towards} possess extensive medical knowledge and strong medical reasoning abilities, as evidenced by their outstanding performance in medical question-answering (MQA) tasks and medical licensing examinations.

When researchers set out to develop LLM-based assisted clinical interview systems, the first challenge they encounter is the scarcity of interview dialogue data, primarily due to privacy concerns. In response, some researchers turn to multi-turn medical dialogues collected from online health consultation platforms to train LLMs~\citep{yang2024zhongjing,chen2023bianque}. However, these dialogues—typically brief interactions of no more than five turns and often containing noise—yield LLMs with limited interview capabilities. Other researchers attempt to synthesize interview dialogues~\citep{liao2023automatic,chen2023soulchat,zhang2023huatuogpt}. They use plain role-play prompts to guide LLMs in transforming source materials, like clinical notes and single-turn health consultations, into multi-turn dialogues. At first glance, such dialogues mirror realistic interview scenarios; however, they harbor a fundamental flaw: their rigid adherence to source materials results in interviews conducted from an 'omniscient perspective,' markedly diverging from the exploratory character of clinical interviews.

Additionally, there is no widely accepted method for assessing the interview capabilities of LLMs in this field. Current evaluation methods are diverse~\citep{zhang2023huatuogpt,chen2023bianque,wang2024notechat}, covering expert evaluations, traditional automatic assessments, and the recently emerging LLM-based automated evaluation. The evaluation metrics they adopt also vary significantly: some are task-specific, some follow the evaluation metrics for natural language generation (NLG) tasks, and some are adapted from evaluation criteria used to assess real-world physicians' interviewing performance—which, unfortunately, are not detailed or comprehensive enough.

In this paper, we propose CliniChat, a framework that integrates multi-source interview knowledge to enable LLMs to simulate real-world clinical interviews. It consists of two modules: Clini-Recon, a method for reconstructing interview dialogues, and Clini-Eval, an LLM-based automated evaluation approach. By incorporating interview knowledge from patient interview guidelines, LLMs, and physicians, Clini-Recon enables LLMs to convert clinical notes into standardized, professional, and empathetic interview dialogues. Clini-Eval features a comprehensive evaluation metric system focused on interview capabilities, coupled with a two-phase evaluation approach called Demo2Eval. Through the seamless integration of both, Clini-Eval allows LLMs to evaluate interview performance like experts. For an overview of the CliniChat framework, please refer to Figure~\ref{fig1}. Experimental results validate CliniChat's effectiveness as a promising solution for LLM-assisted clinical interviews.
%Clini-Eval introduces Demo2Eval, a two-phase evaluation approach inspired by "demonstration teaching," along with a comprehensive system of evaluation metrics that integrates real-world interview scoring criteria, the MVME criteria~\cite{fan2024ai}, and the stylistic differences between LLM-simulated and human physicians. Together, this solution enables LLMs to assess interview performance like experts. 

% With Clini-Recon, we created MedQA-Dialog, a dataset comprising 10,263 highly realistic interview dialogues. By fine-tuning ChatGLM2-6B~\cite{glm2024chatglm} on MedQA-Dialog using P-tuning technology, We developed CliniChatGLM. Experimental results demonstrate a comprehensive upgrade in CliniChatGLM's interview capabilities. Especially in history taking, CliniChatGLM outperforms the GLM-4-Air-simulated physician by 32.9\% on relevant metrics, achieving state-of-the-art (SOTA) performance. 

Contributions of this paper are as follows:% These findings underscore the immense potential of CliniChat in advancing the application of LLMs for assisted clinical interviews. 

\begin{itemize}
    \item We introduce CliniChat, a framework that advances the application of LLMs in clinical interviews. The framework comprises a dialogue reconstruction module and an automated evaluation module, covering the complete pipeline from data construction and model training to evaluation. To the best of our knowledge, CliniChat is the first comprehensive, cost-effective, and efficient solution for integrating LLMs into clinical interviews.
    \item With Clini-Recon, we constructed MedQA-Dialog, a dataset comprising 10,263 highly realistic interview dialogues that span 3,154 diseases across 19 hospital departments. By fine-tuning ChatGLM2-6B~\cite{glm2024chatglm} on MedQA-Dialog, we developed CliniChatGLM, a model specifically for clinical interviews. The dataset, model, and code will be made public upon acceptance of the paper.
    \item We conducted extensive experiments with Clini-Recon, including intrinsic evaluations of interview dialogue quality and extrinsic assessments of interview performance in LLMs. The results show a comprehensive upgrade in CliniChatGLM's interview capabilities. Especially in history taking, it surpasses the GLM-4-Air-simulated physician by 32.9\% on pertinent metrics, achieving SOTA performance.
\end{itemize}

%Clini-Recon is an interview dialogue reconstruction method conditioned on clinical notes. By incorporating multi-source clinical knowledge, it enables LLMs to convert clinical notes into dialogues that closely emulate real-world clinical interviews.
%CliniChat-Eval introduces a LLM-based automated interview evaluation solution. By blending mi-source interview criultteria as our metrics and employing a unique two-phase evaluation approach, Demo2Eval—grounded in "demonstration teaching" principles—CliniChat-Eval enables LLMs like GPT-4o to evaluate interview performance like experts.

\section{Related work}
\label{sec:2}
\paragraph{Models Capable of Multi-Turn Medical Consultations}Super-large language models, whether closed-source~\citep{achiam2023gpt, erniebot2024, glm2024chatglm} or open-source~\citep{touvron2023llama}, and whether general-purpose or specialized for medical use~\citep{singhal2023towards}, show significant potential in assisted clinical interviews. When prompted appropriately, they can partly simulate multi-turn physician-patient interactions~\citep{fan2024ai}. Large multimodal models~\citep{tu2024towards, saab2024capabilities} further expand this potential by integrating medical image analysis and genomic variant detection into medical consultations. However, their utility in real-world clinical interviews remains unexplored. In contrast, LLMs with deployable sizes show weaker potential for clinical interviews, especially for base models like Llama 2-7B~\citep{touvron2023llama} and ChatGLM2-6B. Encouragingly, the Chinese medicine domain has recently seen a boost in deployable models capable of multi-turn health consultations, including BianQue~\citep{chen2023bianque}, ZhongJing~\citep{yang2024zhongjing}, and HuatuoGPT~\citep{zhang2023huatuogpt}. Nevertheless, their interview capabilities remain constrained, largely due to the substantial gap between their fine-tuning datasets and authentic clinical interviews. 
%Med-PaLM C: Towards generalist biomedical AI

\paragraph{Quality-Enhanced Multi-Turn Medical Dialogues}
As data quality largely determines model training effects, researchers have embarked on enhancing the quality of multi-turn medical dialogues. For real-world dialogues, a common approach is to leverage ChatGPT to refine physicians' responses, improving their uniformity, professionalism, and empathy~\citep{chen2023bianque,bao2023disc,yang2024zhongjing}. Whereas for synthetic dialogues, methods are more diverse: \citet{hu2024psycollm} employed a three-step pipeline that incorporates prompts for dialogue generation, evidence evaluation, and refinement; \citet{zhang-etal-2024-cpsycoun} introduced a two-phase framework, Memo2Demo, which builds two roles: a psychological supervisor for consultation note generation and a psychological counselor for dialogue construction; \citet{wang2024notechat} proposed NoteChat, a cooperative multi-agent framework that utilizes LLMs for dialogue planning, role-playing, and polishing. While progress has been made, the dialogues generated by these methods still deviate from real-world clinical interviews, primarily due to their over-reliance on source materials and a narrow focus on isolated skills involved in clinical interviewing.

\section{CliniChat}
\label{sec:3}

\subsection{Source Data}
\label{sec:3.1}
Clinical notes constitute a primary source for reconstructing clinical interview dialogues. They document crucial diagnostic and treatment information from patient encounters, typically organized according to the standardized Subjective, Objective, Assessment, and Plan (SOAP) format~\cite{pearce2016essential}. Nevertheless, privacy concerns significantly impede access to real clinical notes. As an alternative, we utilize the public MQA dataset, MedQA-USMLE~\cite{jin2021disease}, which offers clinical note-like data suitable for research purposes.

\begin{figure*}[t]
  \includegraphics[width=\textwidth]{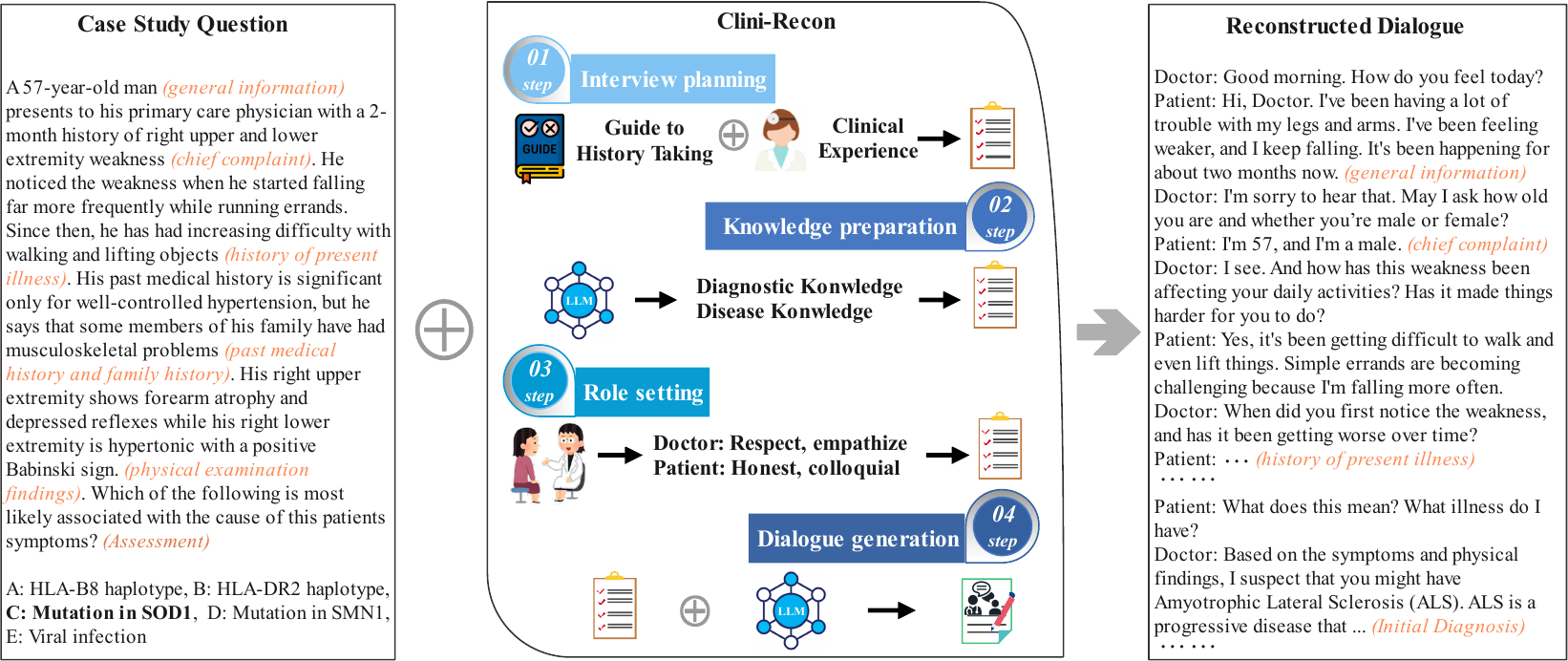}
  \caption{An overview of Clini-Recon, the clinical interview dialogue reconstruction method.}
  \label{fig2}
\end{figure*}

In MedQA-USMLE, case study questions constitute up to 90\%. These questions simulate realistic clinical scenarios by presenting patient cases and examining medical students' patient-centered skills, while following a structure parallel to the SOAP format used in clinical notes. Specifically, they begin with a detailed description of the patient's condition (Subjective), including basic information, chief complaint, history of present illness, past medical history, review of systems, personal history, family history, social history; follow with physical examination and other medical test findings (Objective); and end by asking for either the most likely diagnosis (Assessment) or the most appropriate follow-up examination or treatment (Plan). As single-choice questions, the definitive correct answers provide certainty in both assessment and plan. These features make these case study questions effective substitutes for clinical notes. See Figure~\ref{fig2} for an example case study question and and its SOAP structure breakdown.

We performed a statistical analysis of the MedQA-USMLE training set from the perspective of clinical interviews. The set contains 9,123 case study questions (out of a total of 10,178), spanning 3,154 diseases. We then categorized the questions by standard hospital departments. Specifically, each question was mapped to the department most likely to handle the initial patient visit for the described condition, such as Cardiology, Neurology, Pediatrics, Obstetrics and Gynecology, Orthopedics, Urology, and Psychiatry. The results show that the questions span 19 departments, with the distribution shown in Figure~\ref{fig3} of Appendix~\ref{appendix A}.  

\subsection{Clini-Recon}
\label{sec:3.2}
To elucidate the design philosophy of Clini-Recon, we draw an analogy between reconstructing interview dialogues and preparing a dish. As is well known, preparing a dish requires ingredients (main ingredients, supporting ingredients, and seasonings), directions, and the cooking process. Returning to reconstructing interview dialogues, now all we have are the clinical notes (analogous to main ingredients), so Clini-Recon serves to supplement the missing elements. As illustrated in Figure~\ref{fig2}, we divide it into four sub-tasks: 1) Interview planning (analogous to directions), 2) Knowledge preparation (analogous to supporting ingredients), 3) Role setting (analogous to seasonings), and 4) Dialogue generation (analogous to the cooking process).

\paragraph{Interview Planning}Clinical interviewing is a complex process that combines standardization and personalization: it has clear goal orientation and a phased implementation process while also requiring flexibility to cope with specific patient groups or diseases. For most LLMs, planning interview steps is challenging. \citet{valmeekam2023planning} points out that LLMs' planning capabilities are insufficient, which is reflected in their generated interview dialogues, with deviations in medical logic, confusion in physician-patient roles, and frequent emergence of false information (i.e., "hallucinations")~\citep{wang2024notechat}. Instead, we adopted a manual planning strategy. Specifically, we meticulously planned the interview steps under the guidance of authoritative clinical guidelines and experienced physicians. LLMs are positioned solely as "execution tools," tasked with following pre-planned steps to produce high-quality dialogues.

First, we follow the SOAP format (detailed in Section~\ref{sec:3.1}) to plan the interview process. Based on this, we make two key adaptations: an additional subsection titled "customized inquiry" is introduced to the Subjective section, designed to capture the unique conditions of specific patient groups and diseases; additionally, treatment plans are excluded from the Plan section, as our primary focus is on diagnosis. This interview process is clearly well-suited for initial patient consultations. 

Next, we plan the content for each section of the interview process—specifically, identifying the questions physicians should ask, the concerns patients might raise, and how physicians should respond. Among these sections, the Subjective section, i.e., history taking, being the most complex, requires careful planning. We reference the "GUIDE TO HISTORY TAKING AND EXAMINATION" from University College London Medical School\footnote{\url{https://www.ucl.ac.uk/epidemiology-health-care/sites/epidemiology-health-care/files/history-exam.pdf}}, and incorporate valuable insights from five physicians across different departments. For the other sections, the content is planned based on our experiences. Note that some content requires further completion with knowledge from other subtasks; in these cases, we introduce pre-set placeholders. Meanwhile, we interfuse various interview techniques into the content, such as a mix of open-ended and closed-ended questions, non-leading questions, and non-consecutive questioning, to enhance the efficiency, accuracy, and patient experience of the interview. The well-planned interview content is shown in Figure~\ref{fig4} of Appendix~\ref{appendix B}.

%We reference relevant guidelines from authoritative organization, WHO, clinical medical database, ClinicalKey, and educational platform, Medistudents, while incorporating valuable insights from five physicians across different departments.

\paragraph{Knowledge Preparation}
As mentioned earlier, we have pre-set placeholders in the planned interview content, which effectively prevent our simulated interviews from falling into an 'omniscient' perspective. In fact, these placeholders exist due to a mismatch between the planned interview content and the clinical notes, particularly at the level of clinical knowledge. The objective of Subtask 2—Knowledge Preparation—is to bridge this knowledge gap and fill in these placeholders.

The knowledge gap primarily manifests in two sections: In the Subjective section, clinical notes typically document key symptoms and their progression that support a specific diagnosis; however, they often omit contextual details, such as the patient's lifestyle factors and detailed symptom descriptions—information that our interview content specifically seeks to explore; In the Assessment section, MedQA-USMLE case study questions deliberately omit diagnostic and therapeutic information to test candidates' clinical decision-making abilities—truth that our planned interview content aims to uncover through targeted questioning. To bridge this gap, we constructed a diagnostic knowledge system, incorporating elements such as 'Preliminary Diagnosis' (further divided into 'Most Likely Disease' and 'Differential Diagnosis'), 'Diagnostic Basis,' and 'Confirmatory Tests'; along with a disease knowledge system surrounding the 'Most Likely Disease,' which covers aspects like 'Signs and Symptoms,' 'Risk Factors,' and 'Customized Inquiry.' For the complete knowledge systems, please refer to the prompt for this subtask in Figure~\ref{fig5} in Appendix~\ref{appendix B}. We aligned each element in the knowledge systems with pre-set placeholders in the interview content (as shown in Figure~\ref{fig4}).

\paragraph{Role Setting}Role setting imbues the simulated interview dialogues with human-like qualities. We define the roles of both the patient and the physician based on real-world personality traits and the expectations of each party. During patient interviews, physicians are expected to demonstrate humanistic care—listening attentively and showing genuine sympathy and respect for their patients. Patients or their families generally cooperate fully, answering each question honestly, though they often use everyday language to describe their conditions due to limited medical knowledge. Additionally, for questions that extend beyond the scope of the clinical notes, patients should respond with "No" or "Not sure," avoiding the introduction of unsupported or inconsistent information. For the prompt of this subtask, please refer to Figure~\ref{fig6} of Appendix~\ref{appendix B}

\paragraph{Dialogue Generation}All the "ingredients" for reconstructing interview dialogues are in place, and Clini-Recon will guide the LLM in generating the dialogues. When selecting LLMs, generally, the more advanced, the better—though this comes with higher call costs. Fortunately, extensive manual planning has significantly reduced Clini-Recon's reliance on cutting-edge LLMs, such as GPT-4o or Claude 3.5 Sonnet. Instead, less advanced LLMs, like ERNIE Bot 3.5~\cite{erniebot2024} and GLM-4-Air, are sufficient for the task. In this study, we have selected the highly cost-effective GLM-4-Air, whose call cost is only 1/45th that of GPT-4. Guided by Clini-Recon, GLM-4-Air first performs clinical reasoning on the provided clinical notes and outputs the knowledge required for Subtask 2: Knowledge Preparation. It then seamlessly integrates this knowledge, role settings, and clinical notes into the well-planned interview content, generating standardized, professional, and empathetic simulated clinical interview dialogues. 

\subsection{Clini-Eval}%LLM-based Psychological Counseling
\label{sec:3.3}
\paragraph{Evaluation Metrics}Our metrics for evaluating simulated clinical interviews evolve from three sources of criteria. For the Subjective section, we reference the standardized patient interview scoring criteria from Peking Union Medical College and the MASTER INTERVIEW RATING SCALE from Tulane University School of Medicine\footnote{\url{https://www.dmu.edu/wp-content/uploads/Master-Interview-Rating-Scale.pdf}}, and propose two core metrics—"Mastery of Patient Medical History" (with 16 sub-metrics) and "Interviewing Techniques" (with 8 sub-metrics). In addition to these traditional metrics, we introduced new metrics to accommodate the unique features of our simulated interviews: for the newly added "Customized Inquiry" subsection, we added a metric of the same name; given the differences in interview style between LLM-simulated and real physicians, we incorporated novel metrics such as "Max Two Questions per Inquiry" and "Brief and To-the-Point Responses." For the Objective, Assessment, and Plan sections, we drew on the Multi-View Evaluation Criteria~\cite{fan2024ai} and introduced four major metrics on the consistency of examination results, diagnostic results, diagnostic basis, and confirmatory tests. Ultimately, our evaluation metric system comprises six main metrics and thirty sub-metrics. To the best of our knowledge, this is currently the most comprehensive metric system for evaluating LLM-based simulated clinical interviews. For all evaluation metrics, scores, and descriptions, please see Table~\ref{tab6} in Appendix~\ref{appendix C}.      

\paragraph{Demo2Eval}In clinical interview skill training, the demonstration teaching method plays a key role. Students observe experienced physicians conducting interview demonstration, followed by immediate simulation. The clinical instructor then provides feedback on students' performance by comparing it to the demonstration. Inspired by this, we propose an LLM-based two-phase automated evaluation method, named "Demo2Eval." Given a clinical note and a simulated interview dialogue based on it, the LLM first assumes the role of a senior physician to convert the clinical note into an interview demonstration, then shifts to the role of a clinical instructor to evaluate the simulated dialogue by comparing it with the demonstration.

% In clinical interview skill training, the demonstration teaching method plays a key role. Students observe experienced physicians conducting interview demonstration, followed by immediate simulation. The instructor then provides feedback on students' performance by comparing it to the demonstration. Inspired by this, we propose an LLM-based two-stage automated evaluation method, "Demo2Eval," in which the LLM first acts as a senior physician, providing an interview demonstration based on a clinical note, and then takes on the role of an instructor, evaluating the simulated interview dialogue conditioned on the same clinical note by comparing it to the demonstration.

\textbf{Demo Generation} At this phase, we prompt an LLM to play the role of a senior physician and transform the clinical note into an interview demonstration through a two-step process. Step 1: diagnostic conclusion extraction, which asks the LLM to extract diagnostic conclusions from the clinical note, including the "Most Likely Disease," "Differential Diagnoses," "Diagnosis Basis," and "Confirmation Tests." Step 2: history-taking planning, which requires the LLM to provide a detailed history-taking plan based on the diagnostic conclusions and the clinical note. For MedQA-USMLE case study questions, Step 1 differs slightly: instead of directly extracting the diagnostic conclusion, the LLM must reason through it. To ensure high-quality interview demonstrations, we use GPT-4o to simulate the senior physician. The prompt for demo generation is in Figure~\ref{fig7} of Appendix ~\ref{appendix D}.

\textbf{Comparative Evaluation} At this phase, we assign the role of a clinical instructor to an LLM and prompt it to assess the physician's performance in simulated clinical interviews by referencing the interview demonstration. The evaluation process begins with a subjective assessment, where the LLM compares each point in the interview demonstration with the interview dialogue and provides an evaluation. This is followed by a quantitative evaluation, in which the LLM assigns scores for each evaluation metric based on the results of the subjective assessment. Finally, the overall interview performance is determined by combining the results of both evaluations. This step-by-step process faithfully reproduces the rigorous procedure of real-world clinical interview scoring. To ensure the reliability of the evaluation, we use GPT-4o to simulate the examiner. For the comparative evaluation prompt, please refer to Figure~\ref{fig8} in the appendix.

\section{Experiments}
\label{sec:4}
% In this section, we designed and implemented a series of evaluation experiments, ranging from expert evaluation to automated evaluation, and from internal evaluation to external evaluation, to comprehensively validate the effectiveness of CliniChat.
\subsection{MedQA-Dialog}
\label{sec:4.1}

\begin{table}[ht]
  \centering
  \begin{tabular}{lc}
    \hline
    \textbf{Statistical Index} & \textbf{Value} \\
    \hline
         Max dialogue turns&67\\
         Min dialogue turns&19\\
         Avg. dialogue turns&32\\
         Avg. words in a patient utterance&11.7\\
         Avg. words in a physician utterance&14.8\\\hline
  \end{tabular}
  \caption{Statistics of our MedQA-Dialog dataset.}
  \label{tab1}
\end{table}
%-----------------------------------------------------
We used case study questions from the MedQA-USMLE training and development sets to reconstruct clinical interview dialogues. Guided by Clini-Recon, GLM-4-Air generated 10,263 dialogues that closely simulate real-world clinical interviews, resulting in the MedQA-Dialog dataset. Table~\ref{tab1} presents the dataset statistics, and an example dialogue is provided in Figure~\ref{fig9} in the appendix.

\subsection{Intrinsic Evaluation of CliniChat}
\label{sec:4.2}
To show the superiority of our MedQA-Dialog dataset in simulating real-world clinical interviews, we randomly selected 90 dialogues from it for comparative evaluation. Specifically, we compared these with interview dialogues generated by the following methods (for the criteria used to select these comparison methods, please refer to the appendix): 1) Direct role-play prompting + GPT-4o; 2) Direct role-play prompting + GLM-4-Air; 3) Interactive role-play prompting + GLM-4-Air. These methods were applied to the source case study questions corresponding to the 90 dialogues to generate their respective dialogues. Clini-Eval was used to comprehensively evaluate these dialogues. We present the two role-play prompts of the comparison methods in Figure~\ref{fig10} and Figure~\ref{fig11}  of Appendix~\ref{appendix E}. 

%共同影响考虑到对话合成方法和大语言模型对于合成对话质量同时的，for a more 我们的Existing research on reconstructing medical consultation dialogues typically employs two approaches: direct role-play prompting (see Appendix A.1) and interactive role-play prompting (see Appendix A.2). Since these methods need to be used in conjunction with LLMs, we selected GPT-4o and GLM-4-Air. By combining these methods with the LLMs, we established the following three baseline methods: 1) Direct role-play prompting + GPT-4o; 2) Direct role-play prompting + GLM-4-Air; 3) Interactive role-play prompting + GLM-4-Air. Then these methods were applied to the case study questions corresponding to the 90 randomly selected dialogues. Finally, our evaluation method, Clini-Eval, was used to comprehensively evaluate both our dialogues and those synthesized using the baseline methods. The experimental results are presented in Table 2.

The overall experimental results are shown in Table ~\ref{tab2}. Statistical analysis shows that Clini-Recon generates significantly more dialogue turns, better mirroring the natural flow of real clinical interviews and resulting in higher patient satisfaction. Moreover, it maintains concise utterances from both physicians and patients, facilitating better patient understanding and engagement in the conversation. From a clinical perspective, the aggregate interview performance of Clini-Recon surpasses the strongest baseline by 28.9\%. While showing marginally lower scores in the Examination Results Consistency and Diagnostic Results Consistency metrics, it demonstrates remarkable improvements across other metrics. Most impressively, it outperforms the next-best method by 50.6\% in Mastery of Patient Medical History and 22.5\% in Interview Techniques. These results provide strong evidence that Clini-Recon, by incorporating multi-source interview knowledge into GLM-4-Air, significantly improves the quality of reconstructed clinical dialogues, particularly in gathering patient history - a crucial component of clinical interviews.

\begin{table*}[t]
\resizebox{1.0\textwidth}{!}{
    \centering
    \begin{tabular}{lccccccccc}
         \toprule
    \multirow{3}{*}{\textbf{Method}} & \multicolumn{2}{c}{\textbf{Statistical Indices}} & \multicolumn{7}{c}{\textbf{Interview Evaluation Metrics}}\\
    \cmidrule(r){2-3}
      \cmidrule(r){4-10}&\makecell{Avg.\\Turns}&\makecell{Avg. Words \\Phys. / Pt.}&\makecell{Medical\\History}&\makecell{Interview\\Techniques}&\makecell{Medical\\Exam}& \makecell{Diagnosis\\Result}& \makecell{Diagnosis\\Basis}& \makecell{Confirm.\\Tests}& \makecell{Total\\Score}\\  
     \midrule
Direct Role-play + GLM-4-Air & 8.2 & 33.7 / 18.2 & \underline{21.54} & 18.36 & 3.23 &7.93&7.45&3.64&62.15 \\
Direct Role-play + GPT-4o & 10.7& 27.3 / 13.5  & 20.24 & \underline{19.03} & \textbf{3.71} &\textbf{8.83}&\underline{7.51}&\underline{3.86}& \underline{63.18}\\
Interactive Role-play + GLM-4-Air & 7.8 & 48.8 / 26.2 & 16.33 & 14.25 &  2.95 & 7.08 & 6.47 & 2.86&49.94\\
\textbf{Clini-Recon + GLM-4-Air} & 28.7 & 18.5 / 13.1  & \textbf{32.44} & \textbf{23.31} &  \underline{3.52} & \underline{8.79} & \textbf{8.45} & \textbf{4.92}& \textbf{81.43}  \\           
    \bottomrule
    \end{tabular}}
    \caption{Intrinsic evaluation results on CliniChat. The best score is in-bold, while the second best score is underlined.}
    \label{tab2}
\end{table*}

%------------------------------------------------------
\begin{table*}
\resizebox{1.0\textwidth}{!}{
\centering
    \begin{tabular}{lcccccccc}
         \toprule
    \multirow{3}{*}{\textbf{Method}} & \multirow{3}{*}{\textbf{Department}} & \multicolumn{7}{c}{\textbf{Metrics}}\\
      \cmidrule(r){3-9}& &\makecell{Medical\\History}&\makecell{Interview\\Techniques}&\makecell{Medical\\Exam}& \makecell{Diagnosis\\Result}& \makecell{Diagnosis\\Basis}& \makecell{Confirm.\\Tests}& \makecell{Total\\Score}\\  
     \midrule
        \multirow{7}{*}{\makecell{Direct\\Role-play\\+\\GLM-4-Air}}%{Direct Role-play} 
& Cardiology & 20.38 & 18.72 & 3.15 &7.78 & 7.26 & 3.53 & 60.82 \\
& Endocrinology & 21.91 & 19.08 & 3.32 & 8.48 & 7.81 & 4.11 & 64.71\\
& Neurology & 23.92 & 19.26 &  3.23 & 7.67 & 7.33 & 4.00 & 65.41\\
& Infectious Diseases & 21.45 & 18.67 & 3.28 & 8.00 & 7.62 & 3.69 & 62.71\\ 
& Psychiatry & 19.58 & 16.67 &  3.35 & 7.56 & 6.89 & 3.31 & 57.36\\
& Gynecology & 21.12 & 18.69 & 3.14 & 8.10 & 7.52 & 3.60 & 62.17\\
& Pediatrics & 20.83 & 17.42 & 3.34 & 7.43 & 7.05 & 2.91 & 58.98\\
 \midrule
 \multirow{7}{*}{\makecell{\textbf{Clini-Recon}\\\textbf{+}\\\textbf{GLM-4-Air}}}
& Cardiology & 34.81 (+71\%) & 23.87 (+28\%) & 3.57 & 8.89 & 8.59 & 5.20 & 84.93 (+40\%) \\
& Endocrinology & 34.41 (+57\%) & 23.54 (+23\%) & 3.62 & 9.05 & 8.76 & 5.14 & 84.52 (+31\%)\\
& Neurology & 36.23 (+51\%) & 23.67 (+23\%) &  3.45 & 8.78 & 8.33 & 5.70 & 86.16 (+32\%)\\
& Infectious Diseases & 32.8 (+53\%) & 22.64 (+21\%) & 3.60 & 9.33 & 8.67 & 5.31 & 82.35 (+31\%)\\ 
& Psychiatry & 28.57 (+46\%) & 18.97 (+14\%) &  3.50 & 7.89 & 7.78 & 4.21 & 70.92 (+24\%)\\
& Gynecology & 32.61 (+54\%) & 22.04 (+18\%) & 3.42 & 8.29 & 7.81 & 4.71 & 78.88 (+27\%)\\
& Pediatrics & 32.37 (+55\%) & 22.26 (+28\%) & 3.53 & 8.57 & 8.48 & 5.14 & 80.35 (+36\%)\\
    \bottomrule
    \end{tabular}}
        \caption{Intrinsic evaluation results on CliniChat categorized by hospital department. The values in parentheses indicate the metric improvement of our method vs. the baseline method for dialogues from the same department.}%In the last row of the table, we present the percentage improvement of metrics for Memo2Demo compared to role-play method.
        \label{tab3}
\end{table*}

To gain insight into the adaptability of Clini-Recon, we categorized the intrinsic evaluation results by hospital department. Here, we narrow our focus to seven departments with different interview emphases and higher dialogue proportions: Cardiology, Endocrinology, Neurology, Infectious Diseases, Psychiatry, Gynecology, and Pediatrics. The results are presented in Table~\ref{tab3}. 

As shown in Table~\ref{tab3}, Clini-Recon's adaptability varies across departments. It demonstrates the highest adaptability in Cardiology and Neurology, where its reconstructed dialogues exhibit the most pronounced enhancements, with the primary contribution coming from the mastery of patient medical history, achieving impressive improvements of 71\% and 57\%, respectively. This suggests that Clini-Recon aligns well with the interview patterns of Cardiology and Neurology. In contrast, Psychiatry poses the greatest challenge, with considerable but relatively least improvements in both aggregate performance and history taking. This limitation likely arises from the emphasis of psychiatric consultations on psychological symptoms and emotional states, aspects that require more sophisticated interpretation than Clini-Recon currently provides. These differences in adaptability highlight the specificity of clinical interviews across departments and suggest that future versions of Clini-Recon should integrate more nuanced, department-specific interview knowledge to better adapt to various departments.

\subsection{Extrinsic Evaluation of CliniChat}
\label{sec:4.3}
\paragraph{CliniChatGLM} To evaluate how interview dialogues generated by Clini-Recon can enhance the clinical interview capabilities of LLMs, we fine-tuned ChatGLM2-6B\footnote{\url{https://huggingface.co/THUDM/chatglm2-6b}} on the MedQA-Dialog dataset using the P-Tuning v2 technique~\cite{liu2021p}. During fine-tuning, only the physician's utterances were used as training labels. In this way, we developed an LLM tailored for clinical interview tasks, named CliniChatGLM. For hyperparameter setting, see Table~\ref{tab5} in the appendix.

\paragraph{Baselines and Evaluation Dataset}We selected three groups of models as baselines: our backbone model, ChatGLM2-6B; two close rivals to GPT-4o, GLM-4-Air and Spark4.0 Ultra\footnote{\url{https://xinghuo.xfyun.cn/sparkapi}}; and two open-source Chinese medical LLMs, BianQue~\cite{chen2023bianque} and HuatuoGPT~\cite{zhang2023huatuogpt}.

\begin{table*}[!ht]
\resizebox{1.0\textwidth}{!}{
    \centering
    \begin{tabular}{lccccccccc}
         \toprule
    \multirow{3}{*}{\textbf{Model}} & \multicolumn{2}{c}{\textbf{Statistical Indices}} & \multicolumn{7}{c}{\textbf{Interview Evaluation Metrics}}\\
    \cmidrule(r){2-3}
      \cmidrule(r){4-10}&\makecell{Avg.\\Turns}&\makecell{Avg. Words \\Phys. / Pt.}&\makecell{Medical\\History}&\makecell{Interview\\Techniques}&\makecell{Medical\\Exam}& \makecell{Diagnosis\\Result}& \makecell{Diagnosis\\Basis}& \makecell{Confirm.\\Tests}& \makecell{Total\\Score}\\  
     \midrule
BianQue & 7.7 & 12.9 / 33.4 & 11.38 & 15.04 & 1.21 & 2.34 & 2.04 & 1.60 & 33.61 \\
HuatuoGPT & 5.2 & 261.3 / 61.7 & 10.97 & 16.41 & 2.75 & 4.85 & 4.34 & 3.26 & 42.58\\
Spark4.0 Ultra & 9.9 & 157.8 / 33.5 & 19.91 & \underline{18.34} & \underline{2.92} & 5.96 & 5.30 & 3.51 & 55.94\\
ChatGLM2-6B & 11.2 & 78.8 / 31.8 & 15.86 & 16.23 &  1.65 & 2.87 & 2.66 & 2.02 & 41.29\\
GLM-4-Air & 7.0 & 158.4 / 46.2 & \underline{21.72} & 17.67 & \textbf{3.23} & \textbf{7.42} & \textbf{6.91} & \textbf{3.96} & \underline{60.91}\\
\textbf{CliniChatGLM} & 33.1 & 13.8 / 20.2 & \textbf{29.62} & \textbf{22.74} & 2.76 & \underline{6.28} & \underline{5.91} & \underline{3.83}& \textbf{71.14} \\           
    \bottomrule
    \end{tabular}}
    \caption{Extrinsic evaluation results on CliniChat. The best score is in-bold, while the second best is underlined.}
    \label{tab4}
\end{table*}

Comparative experiments were conducted on the MedQA-USMLE test set. Due to the high cost of GPT-4o API calls, we did not use all case study questions in the set for extrinsic evaluation. Instead, we first randomly selected 100 case study questions and manually filtered them based on whether they contained sufficient information about the chief complaint, medical history, and medical examination results. Finally, 70 questions were selected and used as the extrinsic evaluation dataset.

\paragraph{Automatic Evaluation}Given a case study question, we prompt GLM-4-Air to play the patient role and engage in dynamic multi-turn interactions with a physician LLM, and the role setting is consistent with that of Clini-Recon. The interaction process varies depending on the group of physician LLMs: medical LLMs engage directly in the dialogue, while general-purpose LLMs require additional prompts to play the physician role. Dialogues generated from physician-patient LLM interactions are used as subjects for extrinsic evaluation. Clini-Eval is employed to assess the clinical interview capabilities of the physician role in these dialogues. For the detailed role-play prompts, please refer to Figure~\ref{fig12} and Figure~\ref{fig13} in Appendix~\ref{appendix F}. 

\paragraph{Results}The results are presented in Table~\ref{tab4}. Statistical results indicate that CliniChatGLM inherits the characteristics of high interaction turns and concise utterances from MedQA-Dialog, with approximately 28 dialogue turns dedicated to systematic and comprehensive history taking. In contrast, baseline models such as GLM-4-Air, Sark4.0 Ultra, and HuatuoGPT typically make a diagnosis within fewer than five dialogue turns, with longer individual utterances. An analysis of dialogue instances from these baseline models reveals that their lengthy utterances are driven by two primary factors: a tendency to ask multiple questions within a single utterance, and the retention of generalized health advice generation patterns, a characteristic of LLMs trained on single-turn QA tasks.

The Clini-Eval evaluation results reveal CliniChatGLM's exceptional performance in clinical interviewing, particularly in Mastery of Patient Medical History and Interview Techniques, where it surpasses the strongest baseline model by 36.4\% and 28.7\%, respectively. While CliniChatGLM demonstrates substantial improvements over ChatGLM2-6B across the remaining metrics (from left to right 67.3\%, 218.8\%, 222.2\%, and 189.6\%, respectively), it still slightly lags behind GLM-4-Air, particularly in the diagnosis-related metrics. This discrepancy likely stems from GLM-4-Air's more advanced specialized knowledge and clinical reasoning capabilities. These findings provide clear direction for the future development of CliniChatGLM: While continuing to strengthen its patient history-taking capabilities, efforts should also focus on expanding and diversifying the training corpus to enhance its medical knowledge base and clinical reasoning capability.

\section{Conclusion}
\label{sec:conclusion}
In this paper, we present CliniChat, a multi-source knowledge-driven framework that advances the application of LLMs in assisted clinical interviews. The framework consists of two modules: Clini-Recon for interview dialogue reconstruction, and Clini-Eval for simulated interview evaluation, forming an end-to-end pipeline spanning data construction, model training, and evaluation methodologies. Using Clini-Recon, we constructed MedQA-Dialog, a high-quality synthetic interview dialogue dataset. By fine-tuning ChatGLM2-6B on this dataset, we developed CliniChatGLM. Experimental results demonstrate CliniChatGLM's superior performance in simulated clinical interviews, particularly excelling in history-taking compared to other LLMs. In conclusion, CliniChat provides an end-to-end, cost-effective, and efficient solution for LLM-assisted clinical interviews.

\section*{Limitations}
While CliniChat shows great promise in advancing LLM-assisted clinical interviews, several limitations warrant attention beyond those discussed in the Experiments Section. Due to budget and time constraints, state-of-the-art LLMs like GPT-4o were not incorporated into the dialogue reconstruction, leaving uncertainties regarding CliniChat’s full capabilities. Additionally, inherent issues with LLMs, such as knowledge bias and hallucinations, could introduce inaccuracies into the generated interview dialogues, highlighting the need for robust quality validation mechanisms. Our evaluation relied solely on the Clini-Eval-guided GPT-4o automated assessment method, which, while providing a degree of objectivity and accuracy in the evaluation results, cannot fully replace expert clinical judgment. Future work will incorporate expert evaluations to further validate the alignment between automated and human assessments.

\section*{Ethical Statement}

\paragraph{Data Privacy}Although the CliniChat framework is grounded in clinical notes, which inherently raises concerns about privacy disclosure, this study effectively circumvents these issues. We achieve this by using clinical note-like data, specifically the MedQA-USMLE case study questions. The MedQA-USMLE dataset is collected from the United States Medical Licensing Examination and contains no real patient information, ensuring full compliance with HIPAA regulations. Moreover, during the dialogue reconstruction process with Clini-Recon, we relied solely on general medical knowledge and standard interview protocols, excluding any personal patient details. This approach guarantees that the MedQA-Dialog dataset remains in strict compliance with HIPAA regulations.%根据（Qiu等人，2023）制定的数据版权，我们发布了仅供研究使用的多回合对话评估数据集。

\paragraph{Potential Risks of the Model}While the current version of CliniChatGLM captures the 'form' of clinical interviews by successfully replicating doctor-patient interaction patterns, it still falls short of fully achieving the 'essence'. First, being trained exclusively on the synthetic MedQA-Dialog dataset, it will inevitably show poor performance in diagnosing certain specific groups or diseases when the disease or patient groups covered by the dataset is not balanced. In addition, its flexibility, adaptability, and accuracy also cannot match those of human physicians. Second, the absence of reinforcement learning from human feedback may lead to insufficient sensitivity when addressing user privacy concerns. These limitations pose significant medical risks, ranging from potential misdiagnosis to ethical and privacy risks. We emphasize that CliniChatGLM is an early-stage, research-focused model developed to explore the potential of LLMs in assisting clinical interviews, not a solution ready for clinical use. Users should clearly understand that the output of this model is intended solely for research and educational purposes, and all decisions related to diagnosis or treatment must be made by qualified medical professionals.

\bibliography{main}

\begin{thebibliography}{27}
\providecommand{\natexlab}[1]{#1}

\bibitem[{Achiam et~al.(2023)Achiam, Adler, Agarwal, Ahmad, Akkaya, Aleman, Almeida, Altenschmidt, Altman, Anadkat et~al.}]{achiam2023gpt}
Josh Achiam, Steven Adler, Sandhini Agarwal, Lama Ahmad, Ilge Akkaya, Florencia~Leoni Aleman, Diogo Almeida, Janko Altenschmidt, Sam Altman, Shyamal Anadkat, et~al. 2023.
\newblock Gpt-4 technical report.
\newblock \emph{arXiv preprint arXiv:2303.08774}.

\bibitem[{Baidu(2024)}]{erniebot2024}
Baidu. 2024.
\newblock \href {https://yiyan.baidu.com/welcome} {Ernie-bot 4.0}.
\newblock Accessed 5-January-2024.

\bibitem[{Bao et~al.(2023)Bao, Chen, Xiao, Ren, Wu, Zhong, Peng, Huang, and Wei}]{bao2023disc}
Zhijie Bao, Wei Chen, Shengze Xiao, Kuang Ren, Jiaao Wu, Cheng Zhong, Jiajie Peng, Xuanjing Huang, and Zhongyu Wei. 2023.
\newblock Disc-medllm: Bridging general large language models and real-world medical consultation.
\newblock \emph{arXiv preprint arXiv:2308.14346}.

\bibitem[{Butler(2023)}]{butler2023history}
S~Butler. 2023.
\newblock History taking for advanced clinical practitioners: what should you ask.
\newblock \emph{Nursing Times}.

\bibitem[{Chen et~al.(2023{\natexlab{a}})Chen, Wang, Xing, Xu, Fang, Wang, Li, Wu, Liu, Xu et~al.}]{chen2023bianque}
Yirong Chen, Zhenyu Wang, Xiaofen Xing, Zhipei Xu, Kai Fang, Junhong Wang, Sihang Li, Jieling Wu, Qi~Liu, Xiangmin Xu, et~al. 2023{\natexlab{a}}.
\newblock Bianque: Balancing the questioning and suggestion ability of health llms with multi-turn health conversations polished by chatgpt.
\newblock \emph{arXiv preprint arXiv:2310.15896}.

\bibitem[{Chen et~al.(2023{\natexlab{b}})Chen, Xing, Lin, Zheng, Wang, Liu, and Xu}]{chen2023soulchat}
Yirong Chen, Xiaofen Xing, Jingkai Lin, Huimin Zheng, Zhenyu Wang, Qi~Liu, and Xiangmin Xu. 2023{\natexlab{b}}.
\newblock Soulchat: Improving llms’ empathy, listening, and comfort abilities through fine-tuning with multi-turn empathy conversations.
\newblock In \emph{Findings of the Association for Computational Linguistics: EMNLP 2023}, pages 1170--1183.

\bibitem[{Chung and Park(2019)}]{chung2019chatbot}
Kyungyong Chung and Roy~C Park. 2019.
\newblock Chatbot-based heathcare service with a knowledge base for cloud computing.
\newblock \emph{Cluster Computing}, 22:1925--1937.

\bibitem[{Fan et~al.(2024)Fan, Tang, Chen, Wang, Wei, Xi, Huang, and Zhou}]{fan2024ai}
Zhihao Fan, Jialong Tang, Wei Chen, Siyuan Wang, Zhongyu Wei, Jun Xi, Fei Huang, and Jingren Zhou. 2024.
\newblock Ai hospital: Interactive evaluation and collaboration of llms as intern doctors for clinical diagnosis.
\newblock \emph{arXiv preprint arXiv:2402.09742}.

\bibitem[{GLM et~al.(2024)GLM, Zeng, Xu, Wang, Zhang, Yin, Zhang, Rojas, Feng, Zhao et~al.}]{glm2024chatglm}
Team GLM, Aohan Zeng, Bin Xu, Bowen Wang, Chenhui Zhang, Da~Yin, Dan Zhang, Diego Rojas, Guanyu Feng, Hanlin Zhao, et~al. 2024.
\newblock Chatglm: A family of large language models from glm-130b to glm-4 all tools.
\newblock \emph{arXiv preprint arXiv:2406.12793}.

\bibitem[{Hu et~al.(2024)Hu, Dong, Gang, Ma, Zou, Sun, Guo, Yang, and Wang}]{hu2024psycollm}
Jinpeng Hu, Tengteng Dong, Luo Gang, Hui Ma, Peng Zou, Xiao Sun, Dan Guo, Xun Yang, and Meng Wang. 2024.
\newblock Psycollm: Enhancing llm for psychological understanding and evaluation.
\newblock \emph{IEEE Transactions on Computational Social Systems}.

\bibitem[{Hwang et~al.(2020)Hwang, Lee, Hyun, and Lee}]{9289621}
Tae-Ho Hwang, JuHui Lee, Se-Min Hyun, and KangYoon Lee. 2020.
\newblock \href {https://doi.org/10.1109/ICTC49870.2020.9289621} {Implementation of interactive healthcare advisor model using chatbot and visualization}.
\newblock In \emph{2020 International Conference on Information and Communication Technology Convergence (ICTC)}, pages 452--455.

\bibitem[{Jin et~al.(2021)Jin, Pan, Oufattole, Weng, Fang, and Szolovits}]{jin2021disease}
Di~Jin, Eileen Pan, Nassim Oufattole, Wei-Hung Weng, Hanyi Fang, and Peter Szolovits. 2021.
\newblock What disease does this patient have? a large-scale open domain question answering dataset from medical exams.
\newblock \emph{Applied Sciences}, 11(14):6421.

\bibitem[{Keifenheim et~al.(2015)Keifenheim, Teufel, Ip, Speiser, Leehr, Zipfel, and Herrmann-Werner}]{keifenheim2015teaching}
Katharina~E Keifenheim, Martin Teufel, Julianne Ip, Natalie Speiser, Elisabeth~J Leehr, Stephan Zipfel, and Anne Herrmann-Werner. 2015.
\newblock Teaching history taking to medical students: a systematic review.
\newblock \emph{BMC medical education}, 15:1--12.

\bibitem[{Liao et~al.(2023)Liao, Meng, Liu, Wang, and Wang}]{liao2023automatic}
Yusheng Liao, Yutong Meng, Hongcheng Liu, Yanfeng Wang, and Yu~Wang. 2023.
\newblock An automatic evaluation framework for multi-turn medical consultations capabilities of large language models.
\newblock \emph{arXiv preprint arXiv:2309.02077}.

\bibitem[{Liu et~al.(2021)Liu, Ji, Fu, Tam, Du, Yang, and Tang}]{liu2021p}
Xiao Liu, Kaixuan Ji, Yicheng Fu, Weng~Lam Tam, Zhengxiao Du, Zhilin Yang, and Jie Tang. 2021.
\newblock P-tuning v2: Prompt tuning can be comparable to fine-tuning universally across scales and tasks.
\newblock \emph{arXiv preprint arXiv:2110.07602}.

\bibitem[{Nash(2010)}]{nash2010isabel}
David~B Nash. 2010.
\newblock Isabel, a new diagnostic aid for the 21st century.
\newblock \emph{Pharmacy and Therapeutics}, 35(12):651.

\bibitem[{Pearce et~al.(2016)Pearce, Ferguson, George, and Langford}]{pearce2016essential}
Patricia~F Pearce, Laurie~Anne Ferguson, Gwen~S George, and Cynthia~A Langford. 2016.
\newblock The essential soap note in an ehr age.
\newblock \emph{The Nurse Practitioner}, 41(2):29--36.

\bibitem[{Saab et~al.(2024)Saab, Tu, Weng, Tanno, Stutz, Wulczyn, Zhang, Strother, Park, Vedadi et~al.}]{saab2024capabilities}
Khaled Saab, Tao Tu, Wei-Hung Weng, Ryutaro Tanno, David Stutz, Ellery Wulczyn, Fan Zhang, Tim Strother, Chunjong Park, Elahe Vedadi, et~al. 2024.
\newblock Capabilities of gemini models in medicine.
\newblock \emph{arXiv preprint arXiv:2404.18416}.

\bibitem[{Singhal et~al.(2023)Singhal, Tu, Gottweis, Sayres, Wulczyn, Hou, Clark, Pfohl, Cole-Lewis, Neal et~al.}]{singhal2023towards}
Karan Singhal, Tao Tu, Juraj Gottweis, Rory Sayres, Ellery Wulczyn, Le~Hou, Kevin Clark, Stephen Pfohl, Heather Cole-Lewis, Darlene Neal, et~al. 2023.
\newblock Towards expert-level medical question answering with large language models.
\newblock \emph{arXiv preprint arXiv:2305.09617}.

\bibitem[{Touvron et~al.(2023)Touvron, Martin, Stone, Albert, Almahairi, Babaei, Bashlykov, Batra, Bhargava, Bhosale et~al.}]{touvron2023llama}
Hugo Touvron, Louis Martin, Kevin Stone, Peter Albert, Amjad Almahairi, Yasmine Babaei, Nikolay Bashlykov, Soumya Batra, Prajjwal Bhargava, Shruti Bhosale, et~al. 2023.
\newblock Llama 2: Open foundation and fine-tuned chat models.
\newblock \emph{arXiv preprint arXiv:2307.09288}.

\bibitem[{Tu et~al.(2024)Tu, Azizi, Driess, Schaekermann, Amin, Chang, Carroll, Lau, Tanno, Ktena et~al.}]{tu2024towards}
Tao Tu, Shekoofeh Azizi, Danny Driess, Mike Schaekermann, Mohamed Amin, Pi-Chuan Chang, Andrew Carroll, Charles Lau, Ryutaro Tanno, Ira Ktena, et~al. 2024.
\newblock Towards generalist biomedical ai.
\newblock \emph{NEJM AI}, 1(3):AIoa2300138.

\bibitem[{Valmeekam et~al.(2023)Valmeekam, Marquez, Sreedharan, and Kambhampati}]{valmeekam2023planning}
Karthik Valmeekam, Matthew Marquez, Sarath Sreedharan, and Subbarao Kambhampati. 2023.
\newblock On the planning abilities of large language models-a critical investigation.
\newblock \emph{Advances in Neural Information Processing Systems}, 36:75993--76005.

\bibitem[{Wang et~al.(2024)Wang, Yao, Yang, Zhou, Li, Wang, Xu, and Yu}]{wang2024notechat}
Junda Wang, Zonghai Yao, Zhichao Yang, Huixue Zhou, Rumeng Li, Xun Wang, Yucheng Xu, and Hong Yu. 2024.
\newblock Notechat: a dataset of synthetic patient-physician conversations conditioned on clinical notes.
\newblock In \emph{Findings of the Association for Computational Linguistics ACL 2024}, pages 15183--15201.

\bibitem[{Yang et~al.(2024)Yang, Zhao, Zhu, Zhou, Xu, Jia, and Zan}]{yang2024zhongjing}
Songhua Yang, Hanjie Zhao, Senbin Zhu, Guangyu Zhou, Hongfei Xu, Yuxiang Jia, and Hongying Zan. 2024.
\newblock Zhongjing: Enhancing the chinese medical capabilities of large language model through expert feedback and real-world multi-turn dialogue.
\newblock In \emph{Proceedings of the AAAI Conference on Artificial Intelligence}, volume~38, pages 19368--19376.

\bibitem[{Zhang et~al.(2024)Zhang, Li, Tan, Yang, Zhu, Yang, Zhao, Ye, Li, and Hu}]{zhang-etal-2024-cpsycoun}
Chenhao Zhang, Renhao Li, Minghuan Tan, Min Yang, Jingwei Zhu, Di~Yang, Jiahao Zhao, Guancheng Ye, Chengming Li, and Xiping Hu. 2024.
\newblock \href {https://doi.org/10.18653/v1/2024.findings-acl.830} {{CP}sy{C}oun: A report-based multi-turn dialogue reconstruction and evaluation framework for {C}hinese psychological counseling}.
\newblock In \emph{Findings of the Association for Computational Linguistics: ACL 2024}, pages 13947--13966, Bangkok, Thailand. Association for Computational Linguistics.

\bibitem[{Zhang et~al.(2023)Zhang, Chen, Jiang, Yu, Chen, Chen, Li, Wu, Zhiyi, Xiao et~al.}]{zhang2023huatuogpt}
Hongbo Zhang, Junying Chen, Feng Jiang, Fei Yu, Zhihong Chen, Guiming Chen, Jianquan Li, Xiangbo Wu, Zhang Zhiyi, Qingying Xiao, et~al. 2023.
\newblock Huatuogpt, towards taming language model to be a doctor.
\newblock In \emph{Findings of the Association for Computational Linguistics: EMNLP 2023}, pages 10859--10885.

\bibitem[{Zi-xuan et~al.(2023)Zi-xuan, Jin, Chun, Feng-ying, Xing, and Ruo-jia}]{CHENZi-xuan2023MedicineandPhilosophy}
CHEN Zi-xuan, WANG Jin, PENG Chun, GUO Feng-ying, Zhai Xing, and WANG Ruo-jia. 2023.
\newblock \href {https://doi.org/10.12014/j.issn.1002-0772.2023.10.17} {Research on doctor-patient interaction model in online medical consultation platforms}.
\newblock \emph{Medicine and Philosophy}, 44(10):76--80.

\end{thebibliography}

\appendix

\section{Statistics of Case Study Questions}
\label{appendix A}
\begin{figure*}[ht]
  \includegraphics[width=\textwidth]{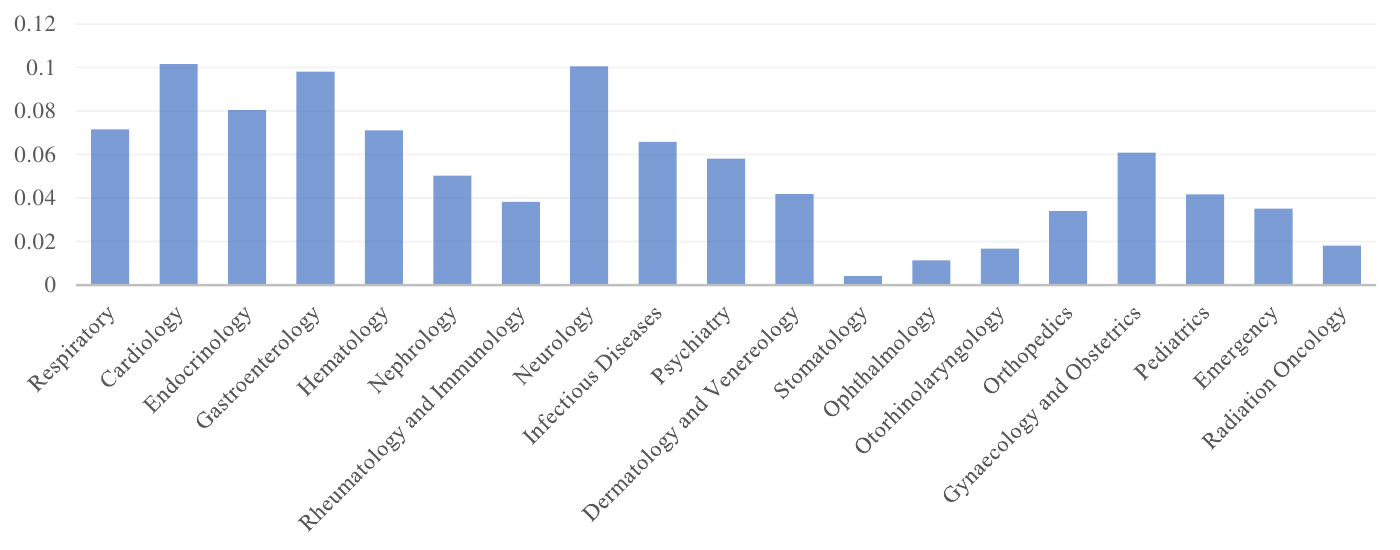}
  \caption{Distribution of case study questions in the MedQA-USMLE training set by hospital departments.}
  \label{fig3}
\end{figure*}

We categorize the case study questions in the MedQA-USMLE training set by standard hospital departments. More specifically, each question was mapped to the department most likely to handle the initial patient visit for the described condition, such as Cardiology, Neurology, Pediatrics, Obstetrics and Gynecology, Orthopedics, Urology, and Psychiatry. The statistical results show that the questions span 19 departments, with the distribution shown in Figure~\ref{fig3}.

\section{Prompts of Clini-Recon}
\label{appendix B}
\begin{figure*}[t]
  \includegraphics[width=\textwidth]{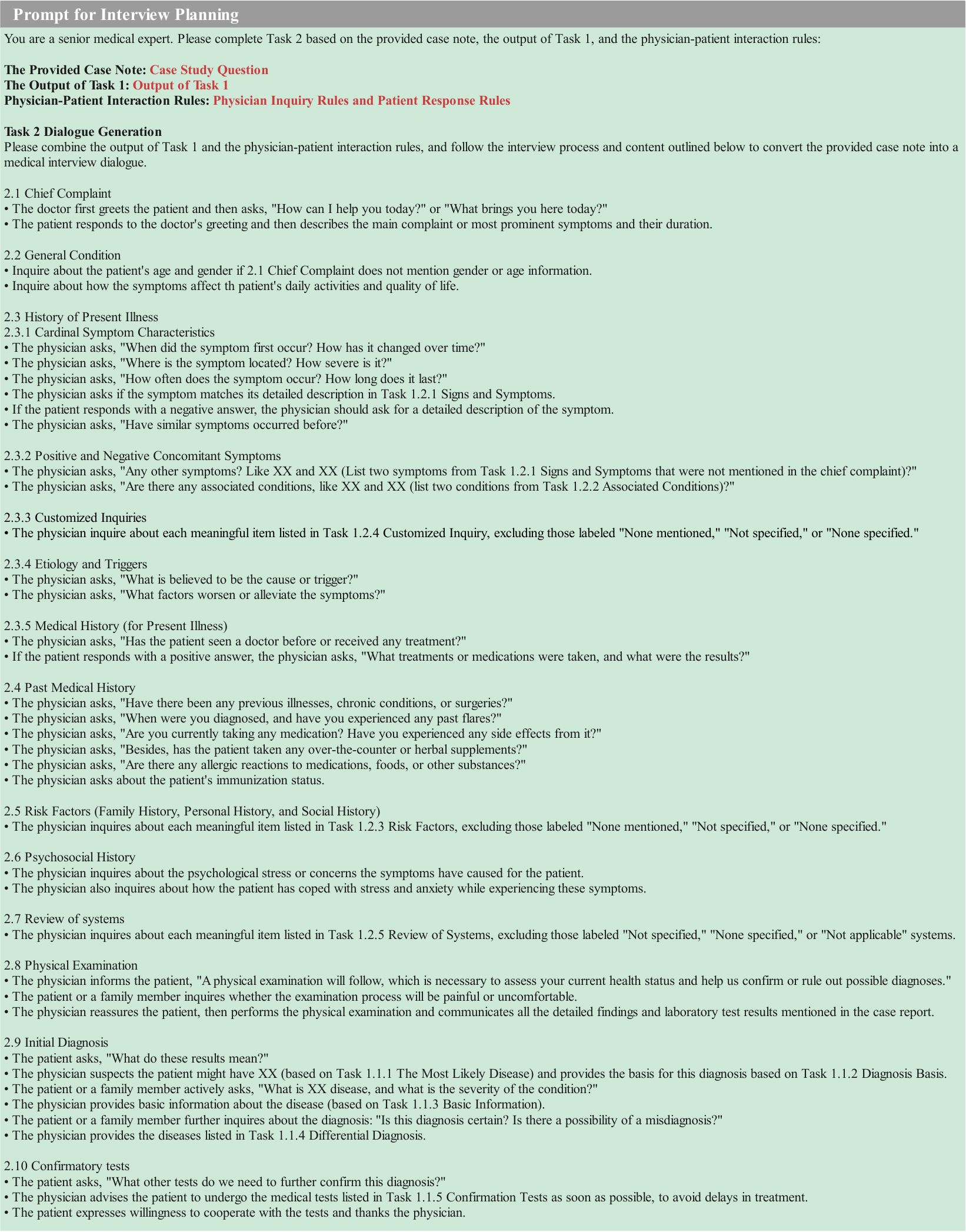}
  \caption{Prompt for the Subtask of Interview Planning.}
  \label{fig4}
\end{figure*}

\paragraph{Prompt for Interview Planning}In Figure~\ref{fig4}, we present the manually planned interview content, i.e., the prompt for interview planning. The secondary headings, such as 2.1 and 2.2, outline the interview process, while the lower-level headings and bullet points provide detailed content. This content integrates knowledge from patient interview guidelines and physicians' expertise, while also incorporating various interview techniques.

\begin{figure*}[t]
  \includegraphics[width=\textwidth]{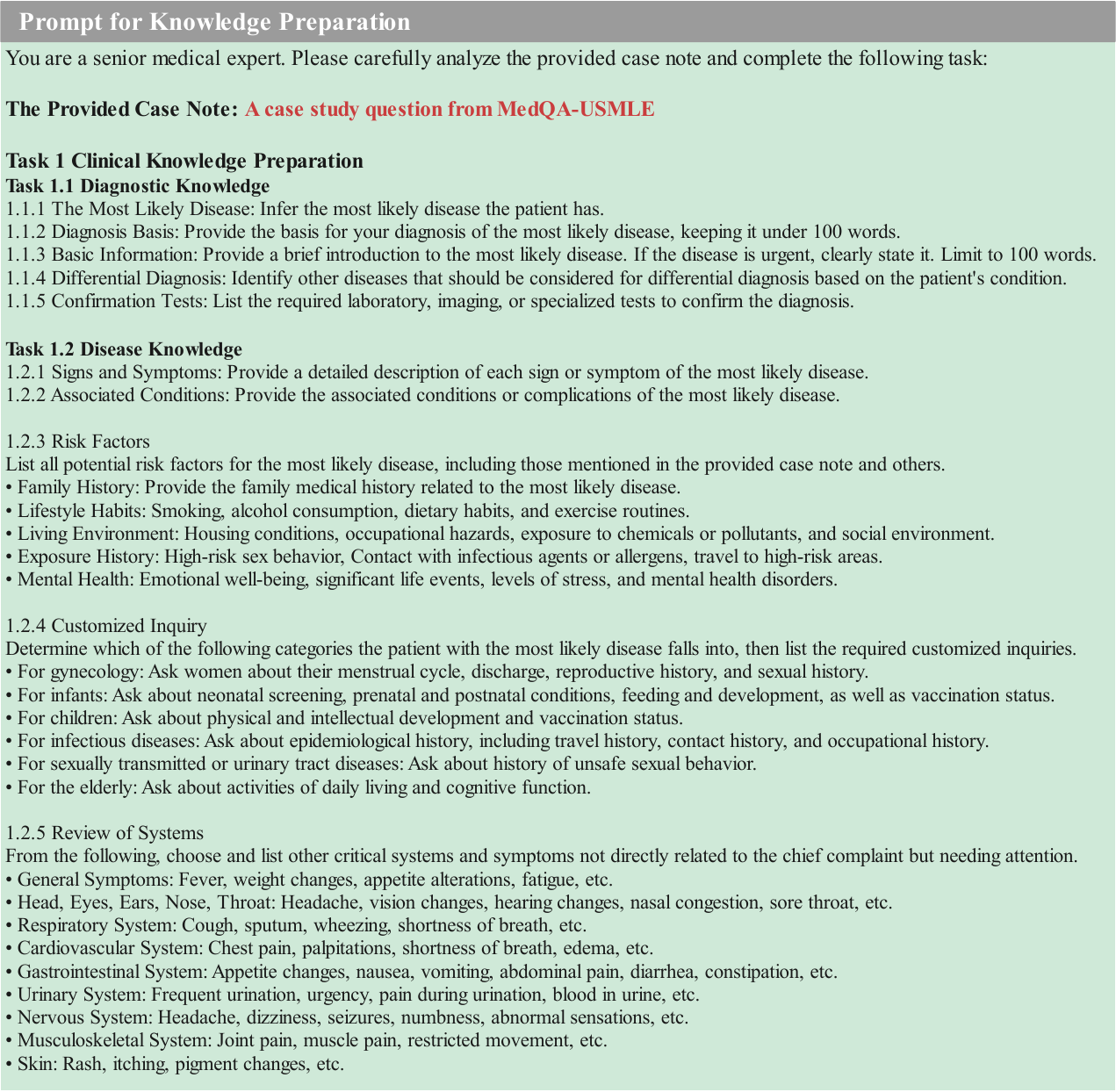}
  \caption{Prompt for the Subtask of Knowledge Preparation.}
  \label{fig5}
\end{figure*}

\paragraph{Prompt for Knowledge Preparation}Figure~\ref{fig5} illustrates the prompt for knowledge preparation, which aims to bridge the clinical knowledge gap between the clinical notes and the planned interview content. Guided by this prompt, LLMs provide the knowledge listed in the diagnostic and disease knowledge systems. This knowledge is then used in the dialogue generation process to fill in the pre-set placeholders within the interview planning prompt.

\begin{figure*}[t]
  \includegraphics[width=\textwidth]{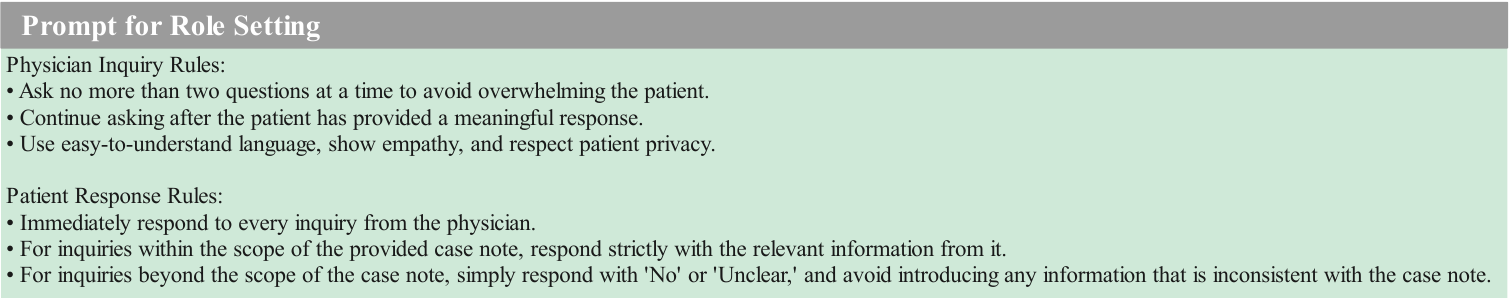}
  \caption{Prompt for the Subtask of Role Setting.}
  \label{fig6}
\end{figure*}

\paragraph{Prompt for Role Setting}In Figure~\ref{fig6}, we present the prompt for the Subtask of Role Setting, which encompasses inquiry rules for physicians and response rules for patients. The physician rules emphasize humanistic care by promoting deep sympathy and respect for patients, whereas the patient rules aim to ensure that responses align with the general patient profile and the information provided in the clinical notes, while maintaining coherent and fluid communication. 

\section{Evaluation Metrics}
\label{appendix C}
To establish a widely accepted metric system for evaluating the physician's performance in LLM-based simulated interview dialogues, we begin with real-world interview scoring criteria, taking into account the differences in interview style between LLM-simulated and real physicians, as well as the Multi-View Evaluation Criteria~\cite{fan2024ai}. We then develop a comprehensive evaluation system comprising six main metrics and thirty sub-metrics. This system covers all sections of the simulated interview dialogues, enabling a thorough evaluation of the LLM-simulated physician's interview performance. For the full list of metrics, scores, and descriptions, please refer to Table~\ref{tab6}.

\section{Prompts of Demo2Eval}
\label{appendix D}
\begin{figure*}[t]
  \includegraphics[width=\textwidth]{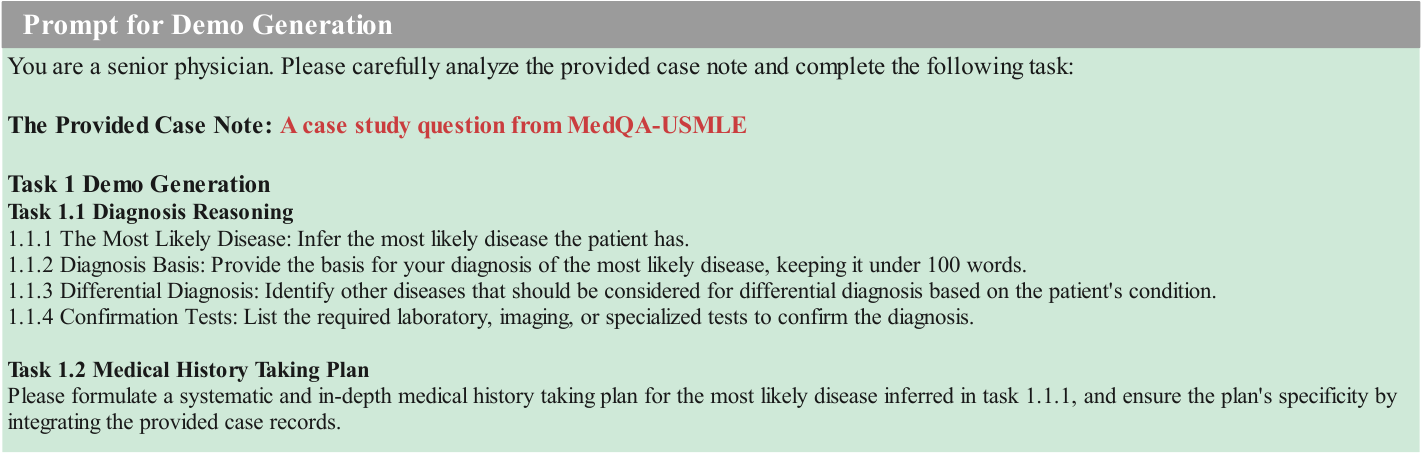}
  \caption{Prompt for the phase of Demo Generation.}
  \label{fig7}
\end{figure*}

\begin{figure*}[t]
  \includegraphics[width=\textwidth]{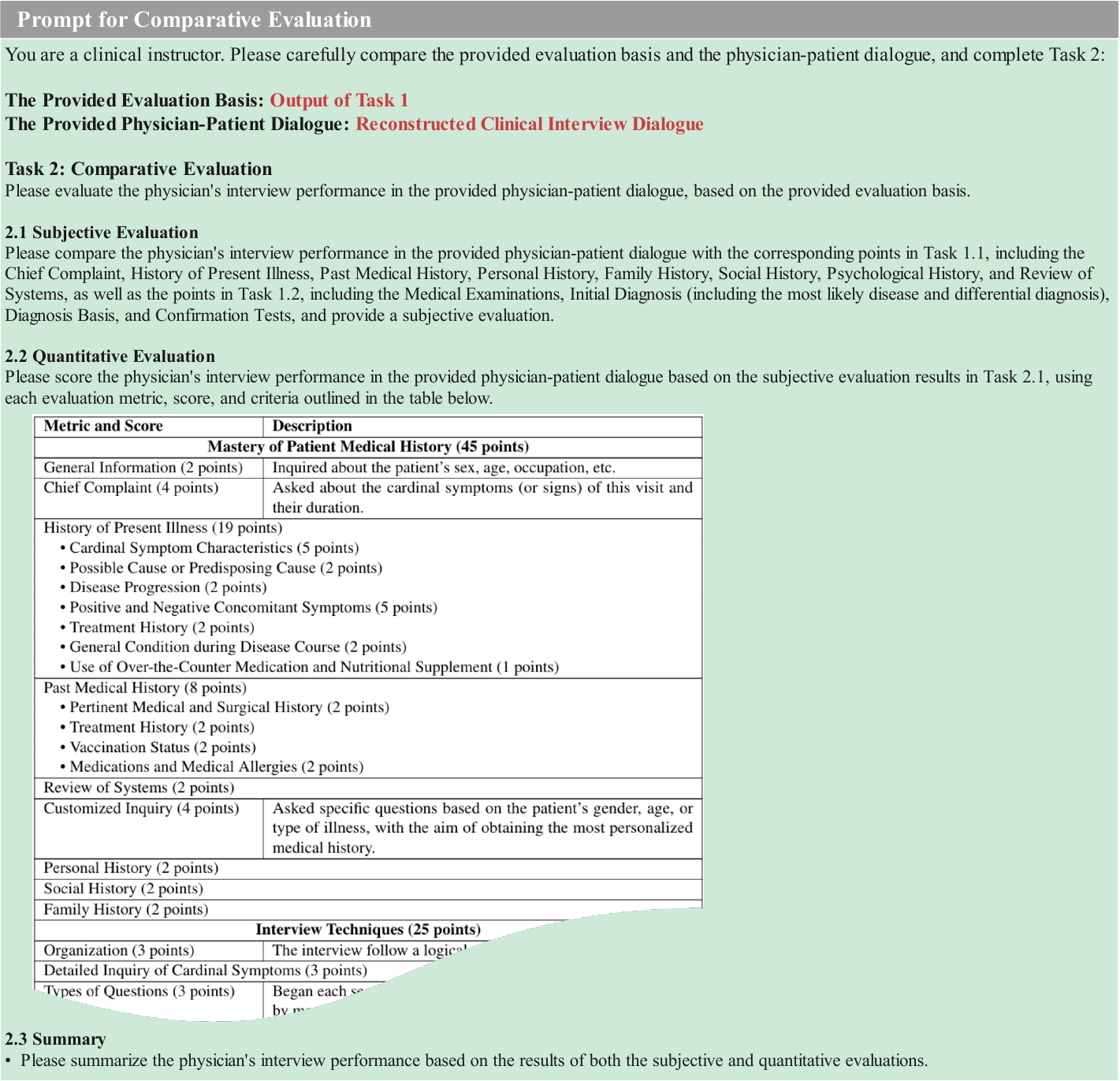}
  \caption{Prompt for the phase of Comparative Evaluation.}
  \label{fig8}
\end{figure*}

Demonstration teaching is fundamental to clinical interview training, where students learn by observing expert physicians, practicing through simulation, and receiving comparative feedback from instructors. Building on this pedagogical model, we introduce Demo2Eval, a two-phase automated evaluation method using LLMs. In the first phase, Demo Generation, the LLM acts as a senior physician to convert the clinical note into an interview demonstration. In the second phase, Comparative Evaluation, the LLM shifts to the role of a clinical instructor to evaluate the simulated dialogue by comparing it to the demonstration. Detailed prompts for both phases are provided in Figure~\ref{fig7} and Figure~\ref{fig8}, respectively.

\section{Intrinsic Evaluation of CliniChat}
\label{appendix E}
\paragraph{Criteria for Selecting comparison Method}It is known that the quality of synthetic clinical interview dialogues is determined by both the prompts and the LLMs they rely on. Existing research on reconstructing medical consultation dialogues typically employs two approaches: direct role-play prompting and interactive role-play prompting. For the LLMs, we selected GPT-4o and GLM-4-Air. By combining these dialogue synthesis methods with the LLMs, we established the following three baseline methods: 1) Direct role-play prompting + GPT-4o; 2) Direct role-play prompting + GLM-4-Air; 3) Interactive role-play prompting + GLM-4-Air. The prompts for both direct role-play and interactive role-play are presented in Figure~\ref{fig10} and Figure~\ref{fig11}.

\section{Extrinsic Evaluation of CliniChat}
\label{appendix F}
\paragraph{Hyperparameter Setting}By fine-tuning ChatGLM2-6B on our MedQA-Dialog dataset using the P-Tuning v2 technique~\cite{liu2021p}, we obtained CliniChatGLM, an LLM specifically designed for interviews. The critical hyperparameters involved in the model training are listed in Table~\ref{tab5}.

\begin{table}[htb]
\centering
\begin{tabular}{ll}
\toprule
\textbf{Hyperparameter} & \textbf{Value} \\
\midrule
Train epochs & 1 \\
Global batch size & 48 \\
Prefix sequence length & 128 \\
Max source length & 2048 \\
Max target length & 128 \\
Learning rate & 1e-2 \\
GPU & 1× NVIDIA V100 \\
\bottomrule
\end{tabular}
\caption{Training hyperparameters}\label{tab5}
\end{table}

\paragraph{Prompts of Extrinsic Evaluation}In the extrinsic evaluation of CliniChat, we have the model under evaluation play the role of a physician conducting a medical interview, while an advanced LLM is prompted to play the patient based on the provided clinical note to cooperate with the physician. The dialogue generated after multiple rounds of interaction serves as the basis for the extrinsic evaluation. Note that general-purpose LLMs, like GLM-4-Air and Spark4.0 Ultra, require additional prompts to effectively assume the role of a physician. Furthermore, since the models under evaluation include both CliniChatGLM, which excels in English, and models like Spark4.0 Ultra, which are more proficient in Chinese, bilingual prompts are provided for both the physician and patient roles, as shown in Figure~\ref{fig12} and Figure~\ref{fig13}.

%###########Table##################
\onecolumn
\begin{longtable}{|p{5cm}|p{10cm}|}
    \hline
    \textbf{Metric and Score} & \textbf{Description} \\
    \hline
    \endfirsthead

    \multicolumn{2}{c}{\tablename\ \thetable{} -- continued from previous page} \\
    \hline
    \textbf{Metric and Score} & \textbf{Description} \\
    \hline
    \endhead

    \hline \multicolumn{2}{r}{Continued on next page} \\
    \endfoot

    \hline
    \endlastfoot

    \multicolumn{2}{|c|}{\textbf{Mastery of Patient Medical History (45 points)}} \\
    \hline
    General Information (2 points) & Inquired about the patient's sex, age, occupation, etc. \\
    \hline
    Chief Complaint (4 points) & Asked about the cardinal symptoms (or signs) of this visit and their duration. \\
    \hline
    \multicolumn{2}{|l|}{History of Present Illness (19 points)} \\
    \multicolumn{2}{|l|}{\quad• Cardinal Symptom Characteristics (5 points)}\\
    \multicolumn{2}{|l|}{\quad• Possible Cause or Predisposing Cause (2 points)}\\
    \multicolumn{2}{|l|}{\quad• Disease Progression (2 points)}\\
    \multicolumn{2}{|l|}{\quad• Positive and Negative Concomitant Symptoms (5 points)}\\
    \multicolumn{2}{|l|}{\quad• Treatment History (2 points)}\\
    \multicolumn{2}{|l|}{\quad• General Condition during Disease Course (2 points)}\\
    \multicolumn{2}{|l|}{\quad• Use of Over-the-Counter Medication and Nutritional Supplement (1 points)} \\
    \hline
    \multicolumn{2}{|l|}{Past Medical History (8 points)}\\
    \multicolumn{2}{|l|}{\quad• Pertinent Medical and Surgical History (2 points)}\\
    \multicolumn{2}{|l|}{\quad• Treatment History (2 points)}\\
    \multicolumn{2}{|l|}{\quad• Vaccination Status (2 points)} \\
    \multicolumn{2}{|l|}{\quad• Medications and Medical Allergies (2 points)} \\
    \hline
    \multicolumn{2}{|l|}{Review of Systems (2 points)} \\
    \hline
    Customized Inquiry (4 points) & Asked specific questions based on the patient's gender, age, or type of illness, with the aim of obtaining the most personalized medical history.\\
    \hline
    \multicolumn{2}{|l|}{Personal History (2 points)} \\
    \hline
    \multicolumn{2}{|l|}{Social History (2 points)} \\
    \hline
    \multicolumn{2}{|l|}{Family History (2 points)} \\
    \hline

    \multicolumn{2}{|c|}{\textbf{Interview Techniques (25 points)}} \\
    \hline
    Organization (3 points) & The interview follow a logical order. \\
    \hline
    \multicolumn{2}{|l|}{Detailed Inquiry of Cardinal Symptoms (3 points)}\\
    \hline
    Types of Questions (3 points) & Began each section with a focused open-ended question followed by more specific questions. \\
    \hline
    Rarely Repetitive Questioning (1 point) & Occasional repetition or duplication solely for clarification or summarization. \\
    \hline
    \multicolumn{2}{|l|}{Non-leading Questions (1 point)} \\
    \hline
    Elicit Patient's Perspective (1 point) & Elicited the patient's perspective on his illness including his beliefs about its beginning, feelings, ideas or cause, function and expectations. \\
    \hline
    Lack of Jargon (2 points) & Used language the patient could easily understand or immediately explained any terminology the patient was not familiar with. \\
    \hline
    Max Two Questions per Inquiry (3 points) & Asked no more than two questions at a time to avoid overwhelming the patient.\\ 
    \hline
    Brief and To the Point Response (3 points) & Responded concisely and accurately, avoiding overly detailed or lengthy responses. \\
    \hline
    Responded Directly (2 points) & Responded align with the patient's concerns and never deviate from the topic.\\
    \hline
    Empathy and Encouragement (2 points) & Expressed understanding, respect, and support for the patient's concerns.\\
    \hline
    Advise urgent care (1 point) & Recommend that the patient seek immediate medical attention.\\
    \hline

    \multicolumn{2}{|c|}{\textbf{Medical Examination and Diagnosis Consistency (30 points)}} \\
    \hline
    Medical Examination Results Consistency (4 points) & Compare the physical examination findings and laboratory test results extracted by the LLM with the interviewer's findings, analyzing their consistency.\\
    \hline
    Diagnosis Consistency (10 points) & Compare the preliminary diagnosis and differential diagnosis results inferred by the LLM with the interviewer's diagnosis results, analyzing their consistency. \\
    \hline
    Diagnostic Basis Consistency (10 points) & Compare the diagnostic basis inferred by the LLM with that of the interviewer and analyze the consistency between them. \\
    \hline
    Confirmatory Tests Consistency (6 points) & Compare the confirmatory test items inferred by the LLM with those recommended by the interviewer, and analyze their consistency. \\
    \hline
    \caption{Evaluation Metrics}\label{tab6}
\end{longtable}

%--------------------------------------------------
%----------------------------------------------------
%-----------------------------------------------------
%----------------------------------------------------
%-----------------------------------------------------
%----------------------------------------------------
\begin{figure*}[t]
  \includegraphics[width=\textwidth]{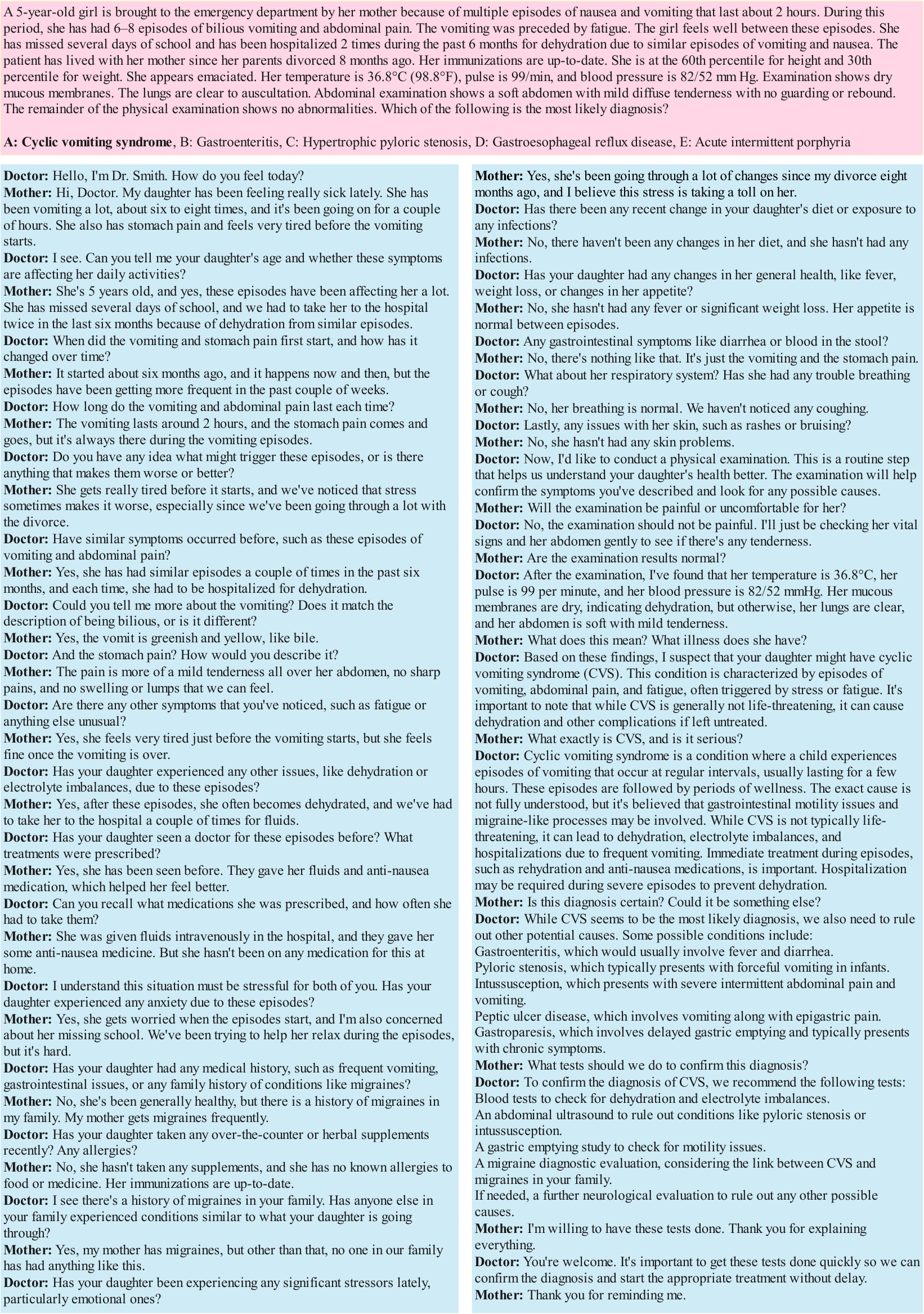}
  \caption{An example dialogue from MedQA-Dialog and its corresponding case study question (top, in pink).}
  \label{fig9}
\end{figure*}

%-------------------------------------------------
\begin{figure*}[h]
  \centering  % 添加这行来居中
  \includegraphics[width=0.55\textwidth]{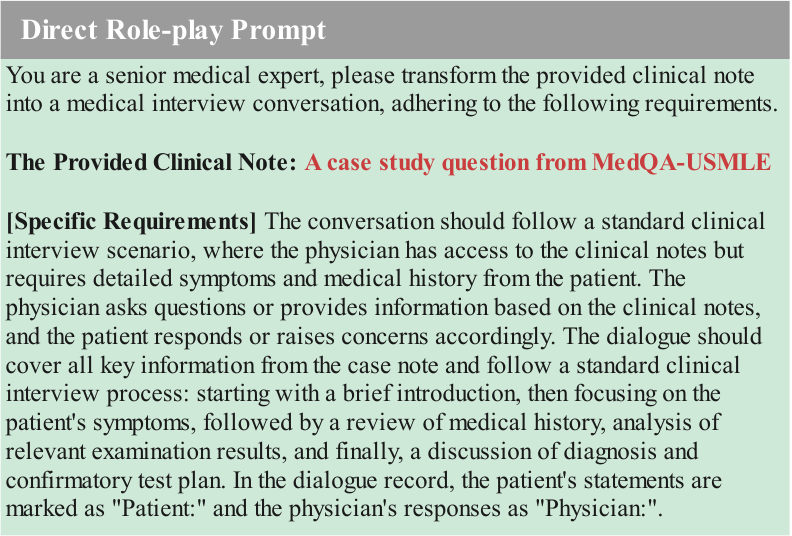}% 把宽度改为0.7倍
  \caption{Direct role-play prompt.}
  \label{fig10}
\end{figure*}

% \begin{figure}[h]
%   \includegraphics[width=\columnwidth]{latex/direct role-play prompt-1.pdf}
%   \caption{Direct role-play prompt.}
%   \label{fig9}
% \end{figure}

%-------------------------------------------------
\begin{figure*}[h]
  \includegraphics[width=\textwidth]{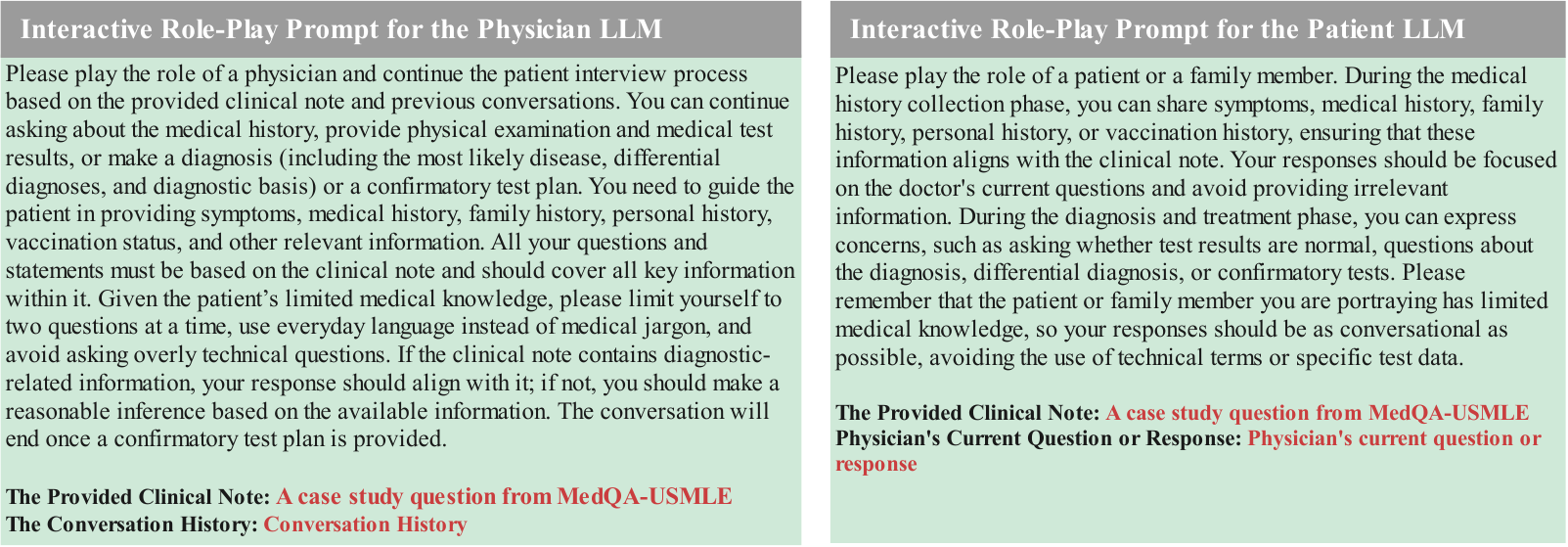}
  \caption{Prompts of interactive role-play.}
  \label{fig11}
\end{figure*}

% \begin{figure}[h]
%   \includegraphics[width=\columnwidth]{latex/interactive role-play-prompts-y-1.pdf}
%   \caption{Prompts of interactive role-play.}
%   \label{fig10}
% \end{figure}

%-----------------------------------------------------
\begin{figure*}[t]
  \includegraphics[width=\textwidth]{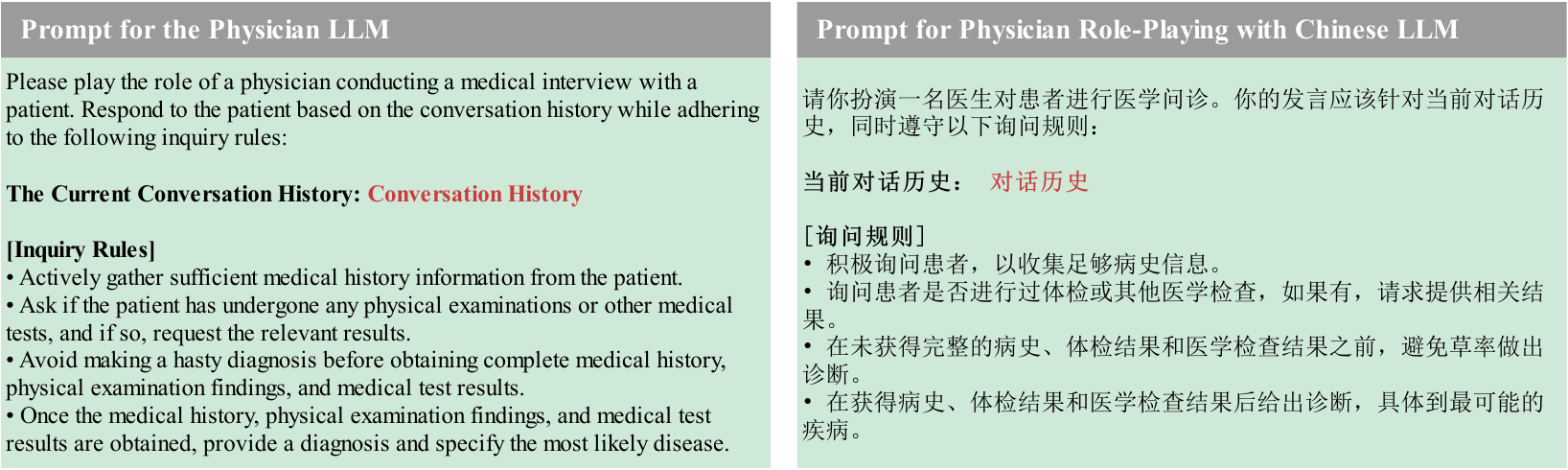}
  \caption{Prompt for the physician LLM in extrinsic evaluation.}
  \label{fig12}
\end{figure*}
%----------------------------------------------------
\begin{figure*}[t]
  \includegraphics[width=\textwidth]{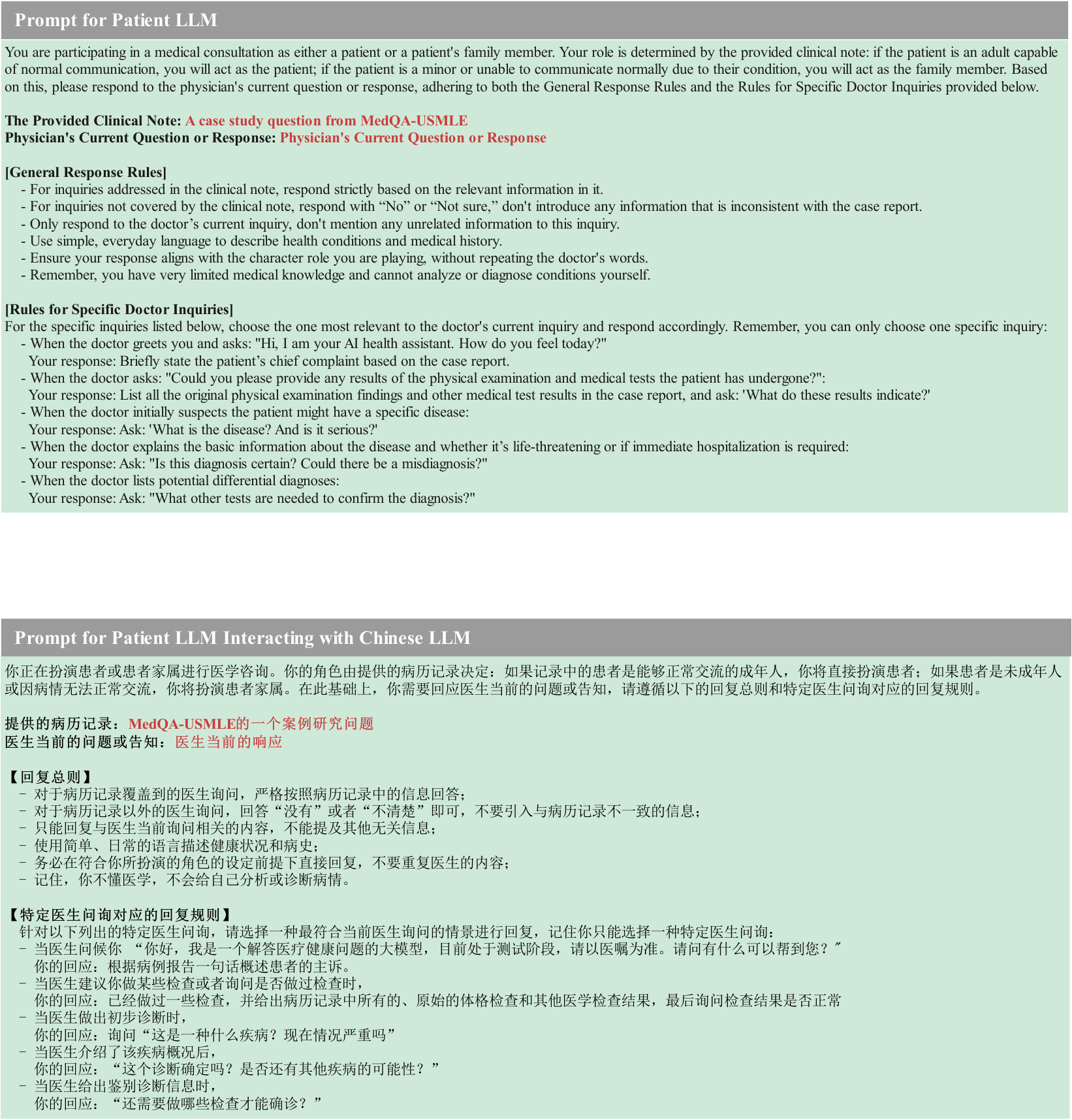}
  \caption{Prompt for the patient LLM in extrinsic evaluation.}
  \label{fig13}
\end{figure*}

\end{document}